\documentclass[sigconf]{acmart}
\acmSubmissionID{2995}
\usepackage{graphicx}
\usepackage{float}
\usepackage{subfig}
\usepackage{multirow}
\usepackage{color}
\AtBeginDocument{%
  \providecommand\BibTeX{{%
    \normalfont B\kern-0.5em{\scshape i\kern-0.25em b}\kern-0.8em\TeX}}}

\setcopyright{acmcopyright}
\copyrightyear{2022}
\acmYear{2022}
\acmDOI{10.1145/3503161.3548394}
\acmConference[MM '22] {Proceedings of the 30th ACM International Conference on Multimedia}{October 10--14, 2022}{Lisboa, Portugal.}
\acmBooktitle{Proceedings of the 30th ACM International Conference on Multimedia (MM '22), October 10--14, 2022, Lisboa, Portugal}

\acmPrice{15.00}
\acmISBN{978-1-4503-9203-7/22/10}
\begin{document}

%\title{Depth Basis for Consistent Structure Estimation in Multi-Camera Systems}
\title{Multi-Camera Collaborative Depth Prediction  via Consistent Structure Estimation}

\author{Jialei Xu}
\affiliation{%
  \institution{Harbin Institute of Technology }
  \city{Harbin}
  \country{China}}

\email{21b903029@stu.hit.edu.cn}

\author{Xianming Liu$^*$}\thanks{$^*$Corresponding author}
\affiliation{%
  \institution{Harbin Institute of Technology \& Peng Cheng Laboratory}
  \city{Harbin}
  \country{China}}
\email{csxm@hit.edu.cn}

\author{Yuanchao Bai}
%\orcid{0000-0002-5694-505X}
\affiliation{%
  \institution{Harbin Institute of Technology }
  \city{Harbin}
  \country{China}
}
\email{yuanchao.bai@hit.edu.cn}

\author{Junjun Jiang}
\affiliation{%
  \institution{Harbin Institute of Technology \& Peng Cheng Laboratory}
  \city{Harbin}
  \country{China}
}
\email{jiangjunjun@hit.edu.cn}

\author{Kaixuan Wang}
\affiliation{%
  \institution{Shenzhen DJI Sciences and Technologies Ltd.}
  \city{Shenzhen}
  \country{China}
}
\email{halfbullet.wang@dji.com}

\author{Xiaozhi Chen}
\affiliation{%
  \institution{Shenzhen DJI Sciences and Technologies Ltd.}
  \city{Shenzhen}
  \country{China}
}
\email{xiaozhi.chen@dji.com}

\author{Xiangyang Ji}
\affiliation{%
  \institution{Tsinghua University}
  \city{Beijing}
  \country{China}
}
\email{xyji@tsinghua.edu.cn}

%-------------------------------------------------------------------

\renewcommand{\shortauthors}{Jialei Xu et al.}

\begin{abstract}

Depth map estimation from images is an important task in robotic systems. Existing methods can be categorized into two groups including multi-view stereo and monocular depth estimation. The former requires cameras to have large overlapping areas and sufficient baseline between cameras, while the latter that processes each image independently can hardly guarantee the structure consistency between cameras. In this paper, we propose a novel multi-camera collaborative depth prediction method that does not require large overlapping areas while maintaining structure consistency between cameras. Specifically, we formulate the depth estimation as a weighted combination of depth basis, in which the weights are updated iteratively by a refinement network driven by the proposed consistency loss. During the iterative update, the results of depth estimation are compared across cameras and the information of overlapping areas is propagated to the whole depth maps with the help of basis formulation. Experimental results on DDAD  and NuScenes datasets demonstrate the superior performance of our method.

\end{abstract}






\maketitle

\section{Introduction}

Generating high-fidelity depth from color is attractive due to that it offers an inexpensive alternative to complement LIDAR sensors. Depth estimation enables agents to reconstruct the surrounding 3D environment, and thus holds great potential applications in self-driving cars, robotics, AR-compositing, etc.

% Benefiting by the cost, size, and dense information, cameras are widely used by robotics, AR, and other applications. Estimating the depth maps of images enables robotics to reconstruct the environment, detect obstacles, which is the corner stone of autonomous navigation (e.g., self-driving, autonomous unmanned vehicles).
Depth estimation without knowing a reference input image is an ill-posed problem. 
To produce dense depth, the current methods typically take as input synchronized stereo pairs or a monocular stream of images, which are referred to as multi-view stereo (MVS)~\cite{yang2020cost,gu2020cascade,watson2021temporal} and monocular depth prediction~\cite{godard2019digging,zhou2017unsupervised,garg2016unsupervised}, respectively. MVS based methods estimate the depth of every pixel using the constrains from multi-view geometry. One pixel has to be observed by multiple cameras distributed in space. The overlap of cameras and baseline between cameras are important for the estimation accuracy. For example, KITTI dataset~\cite{geiger2012we} are collected by a multi-camera systems with a baseline of 0.54m, which limits the application of consuming products. Monocular depth prediction, on the other hand, uses deep learning methods to infer the depth map from only one image. The scale information is unobservable in images such that depth maps of different cameras can hardly be consistent to each other. Inconsistence between cameras will lead to poor reconstruction and downstream tasks.

The modern self-driving cars are usually equipped with multiple cameras to capture the full surround 360$^\circ$ field of view.
This inspires a new line of depth estimation, which works in a multi-camera setting by leveraging cross-camera overlapping contexts. Different from the stereo pair based methods, instead of requiring stereo-rectified or highly-overlapping images, it rather exploits small overlaps (as low as 10\%) between cameras with arbitrary locations~\cite{9712255}. Surprisingly, there are very few works along this way. To the best of our knowledge, there are only two existing works in the literature so far~\cite{9712255,wei2022surround}. Guizilini \textit{et al.} propose the first work along this line, referred to as full surround monodepth (FSM)~\cite{9712255}, which leverages {cross-camera temporal contexts} via {spatio-temporal photometric  constraints} to increase the amount of overlap between cameras. By exploiting known extrinsics between cameras, FSM enforces pose consistency constraints to ensure all cameras follow the same rigid body motion. However, the photometric loss~\cite{godard2019digging} cross cameras used by FSM can be affected brightness change or lens distortion. In addition, it can only utilize the information of the overlapping parts between multiple cameras in the training phase. In the inference stage, images are processed independently and the overlapping parts are ignored. In challenging scenes, the depth consistency will be violated, leading to poor depth estimation accuracy.

In this paper, we propose a novel multi-camera collaborative depth prediction (MCDP) method  that can recover the absolute scale and maintain the structure consistency between multiple cameras, without the requirement of large camera overlapping. 
Specifically, we formulate the depth estimation as a weighted combination of depth basis, in which the weights are updated iteratively by a refinement network driven by the proposed structure consistency loss.
During the iterative update, we align the results of depth estimation across cameras and extract the features of overlapping areas to guide the refinement of the weights, such that the information of small overlapping areas can be propagated to the whole depth maps with the help of basis formulation.
We evaluate the effectiveness of our method on the widely used DDAD~\cite{godard2019digging} and NuScenes~\cite{caesar2020nuscenes} datasets. The contributions of this work can be summarized as follows:

%First, we propose a refinement network driven by the consistency loss to improve the absolute scale and consistency. Depth maps are iteratively projected and compared, while the warped feature maps are used to refine the estimation through the network. Second, since the structure of the depth maps can be well estimated from a single image, multi-view information from overlapping areas provides scale information and detailed refinements, we further reduce the update parameter by depth basis formulation. Each depth estimation is the weighted sum of depth basis predicted by a single image. The weight of different basis is refined by the refinement network, such that small overlapping areas can influence the whole image. 

\begin{figure*}[htb]
\centering
%\subfloat{
%\textcolor{green}{\rule{2.835cm}{1.55mm}}
%\textcolor{cyan}{\rule{2.835cm}{1.535mm}}
%\textcolor{magenta}{\rule{2.835cm}{1.55mm}}
%\textcolor{yellow}{\rule{2.835cm}{1.55mm}}
%\textcolor{red}{\rule{2.835cm}{1.55mm}}
%\textcolor{blue}{\rule{2.835cm}{1.55mm}}
%}
%\vspace{-4.0mm}
%\\
\subfloat{
\includegraphics[width=0.16\linewidth,height=1.5cm]{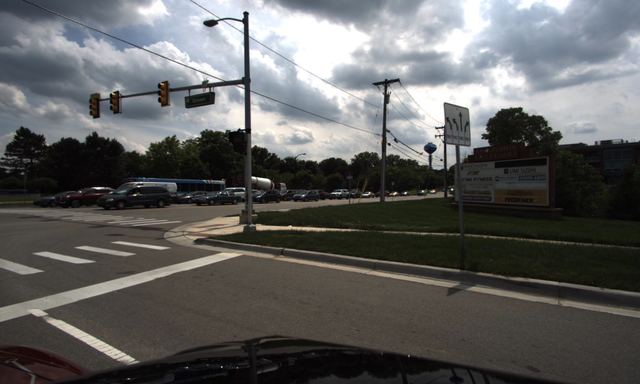}
\includegraphics[width=0.16\linewidth,height=1.5cm]{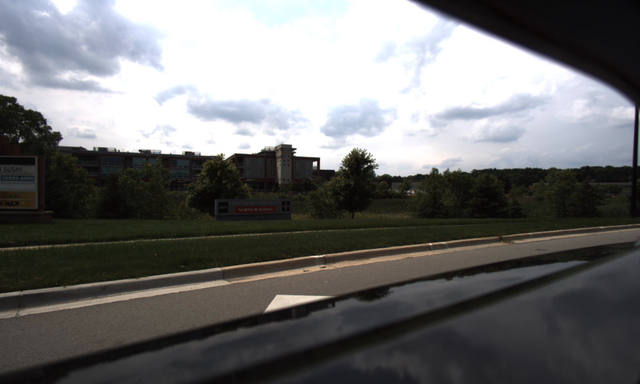}
\includegraphics[width=0.16\linewidth,height=1.5cm]{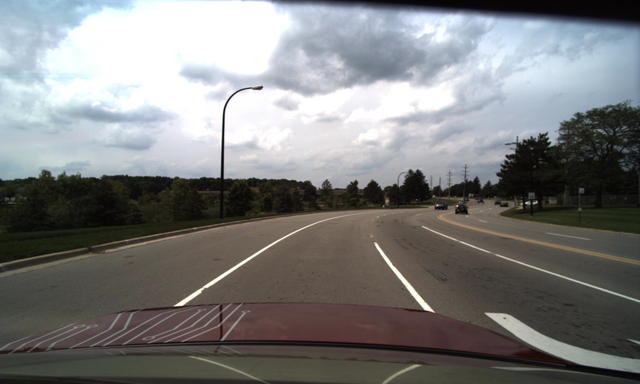}
\includegraphics[width=0.16\linewidth,height=1.5cm]{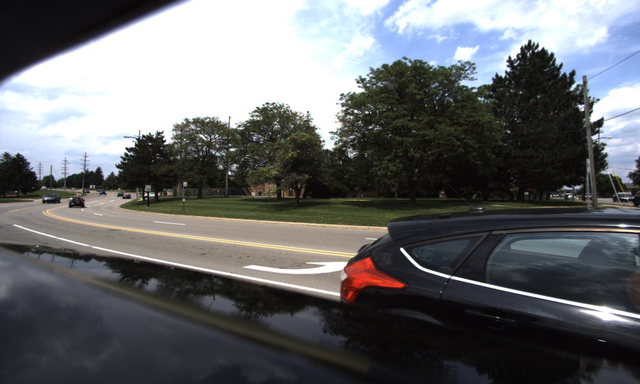}
\includegraphics[width=0.16\linewidth,height=1.5cm]{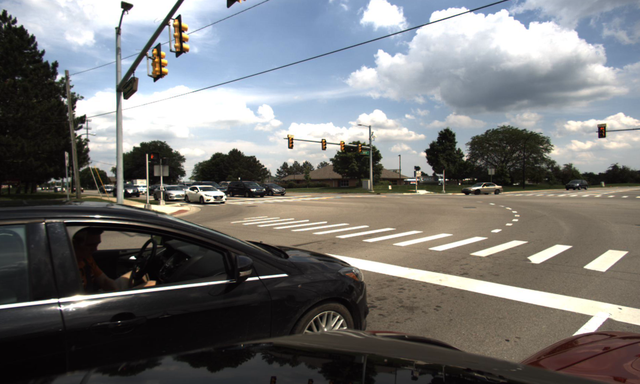}
\includegraphics[width=0.16\linewidth,height=1.5cm]{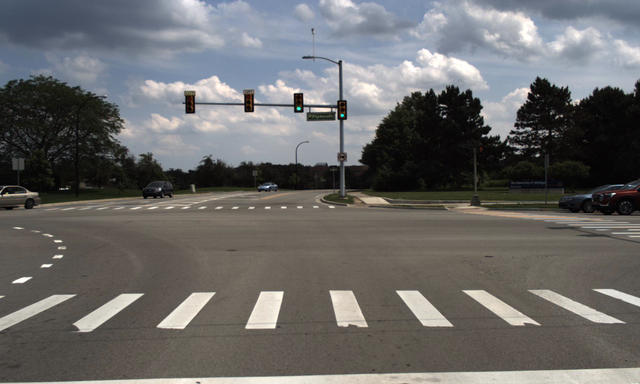}
}
\vspace{-4mm}
\\
\subfloat{
\includegraphics[width=0.16\linewidth,height=1.5cm,height=1.5cm]{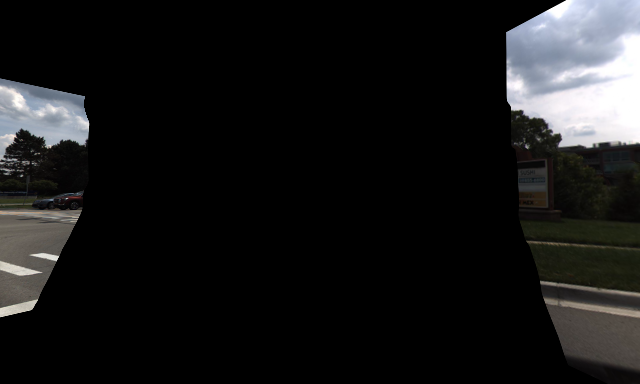}
\includegraphics[width=0.16\linewidth,height=1.5cm,height=1.5cm]{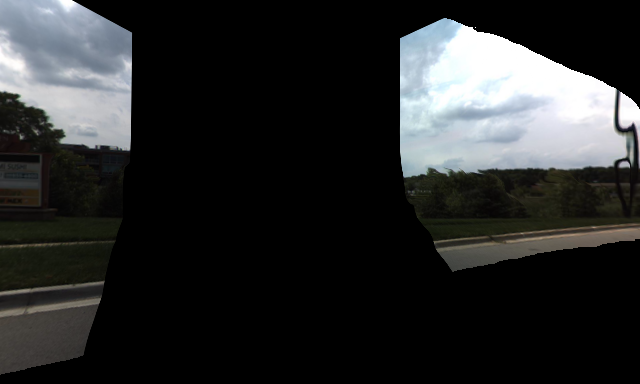}
\includegraphics[width=0.16\linewidth,height=1.5cm,height=1.5cm]{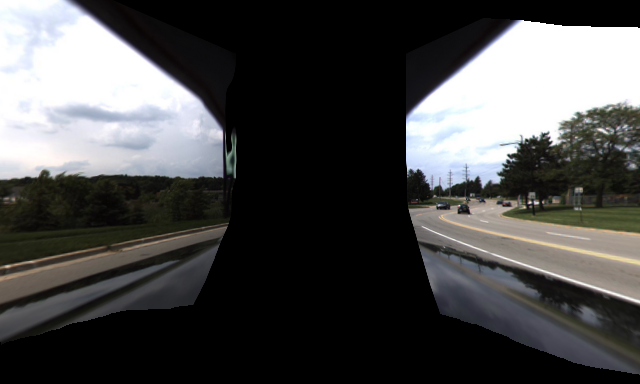}
\includegraphics[width=0.16\linewidth,height=1.5cm,height=1.5cm]{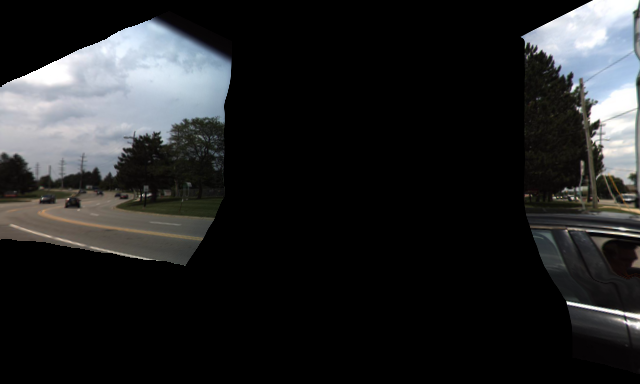}
\includegraphics[width=0.16\linewidth,height=1.5cm,height=1.5cm]{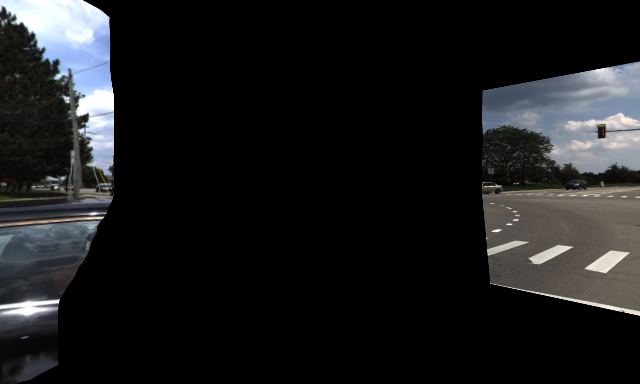}
\includegraphics[width=0.16\linewidth,height=1.5cm,height=1.5cm]{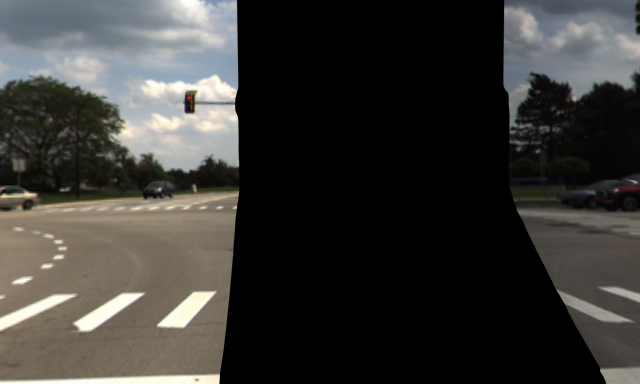}
\vspace{-4mm}
}

\caption{
%\textbf{Overlap area of each camera}
Examples and the overlapping areas of each camera on the DDAD dataset. First row: RGB images of cameras. Second Rows: Synthesized images warped from adjacent cameras.}
% \vspace{-2mm}
%, including surface normals calculated from the predicted depth maps.}
\label{fig:ddad overlap}
\end{figure*}

%\captionsetup[table]{skip=6pt}
%---------------------------------------------------------------------------------------------------------------------------------------------------

\begin{itemize}
\item We introduce a novel multi-camera collaborative depth prediction method that can exploit small overlapping information between cameras to achieve scale and structure consistency of depth predictions.

\item To improve the efficiency and reduce the network parameters, we solve the depth estimation problem through a basis combination scheme, which is referred to as depth basis for consistent structure estimation. The depth basis is estimated for each specific image and the weights are updated by multi-view information. 

%We introduce the consistency of the estimated depth of different cameras in the image overlapping area. And we design the a new loss to restrict the depth in  overlapping part, which we refer to as \textbf{Depth Consistency Loss}.

\item  We demonstrate the effectiveness of the proposed approach on two public multi-camera datasets: DDAD~\cite{godard2019digging} and NuScenes~\cite{caesar2020nuscenes}.
The proposed approach achieves the state-of-the-art performance compared with other approaches.

%Experimental results demonstrate that the proposed DBE achieves promising performance.
\end{itemize}

%\begin{figure}[t!]
%\centering
%\includegraphics[width=1.0\linewidth]{figures/DDAD_dataset/DDAD_Camera.png}
%\caption{ Monodepth  pointcloud in DDAD dataset~\cite{guizilini20203d}}
%\label{fig: DDAD datasets}
%\end{figure}

\section{Related Work}
\subsection{Monocular Depth Estimation}
Relying on the  depth ground-truth, Eigen ~\textit{et al.} ~\cite{eigen2014depth} develop a architecture for coarse-to-fine depth estimation with a scale-invariant loss function. This pioneering work inspires new CNN-based architectures to monodepth estimation~\cite{he2018learning, yin2021virtual, liu2015learning, Wei2021CVPR}. Supervised monocular depth prediction needs to use the ground-truth provided by the depth sensor as the supervision information. In the existing datasets such as NYU~\cite{silberman2012indoor} and KITTI~\cite{geiger2013vision}, depth sensors commonly use stereo cameras and LiDAR to get depth information. With the development of the depth prediction network, supervised depth estimation tries to use classification problem ~\cite{fu2018deep,bhat2021adabins} to predict depth, which has achieved very good results. With the great breakthrough of the transformer network structure in computer vision, ~\cite{ranftl2021vision, xu2021weakly} adopts the model related to the transformer, and have made great progress in many datasets. As a perceptual algorithm, depth prediction need strong geometric constraints. ~\cite{yin2019enforcing,keltjens2021self} use the coplanar constraint of normal vectors and points to strengthen the geometry. 
%Depth prediction is not a completely independent task, and it is often adopted into a multi task model. ~\cite{jiao2018look,mousavian2016joint} combines semantic segmentation and depth prediction. They aim at producing pixelwise information and help each other to improve their predictions at object boundaries.

The drawback of these supervised approaches is that the depth ground truth usually comes from the expensive LiDAR, which must be calibrated and synchronized with the cameras. Moreover, LiDAR depth is sparse compared to available image resolutions. In many outdoor scenes, the surface of objects such as glass and water will affect the range detection of LiDAR. Therefore, the self-supervised depth estimation model receives more and more attention.

~\cite{ozyecsil2017survey} provides a pivotal principle of self-supervised, which has affected the design of many models so far. Their key idea is that obtaining a frame $X_t$ from consecutive ones, and \{$X_{t-1}$, $X_{t+1}$\} can be decomposed into jointly estimating the scene depth for $X_t$ and the camera pose at time $T$ relative to its pose at time t $\pm$ 1. Zhou ~\textit{et al.}~\cite{zhou2017unsupervised} follow this idea. They train the depth network to estimate depth map from the $X_t$ and the  pose network to estimate the pose between the $X_t$ and $X_{t {\pm} 1}$. The photometric loss between $X_t$ and $\hat{X}_{t}$ acts as training loss. Furthermore, Godard ~\textit{et al.}~\cite{godard2019digging} propose a model called Mono2, where the multi-scale approach and per-pixel minimum reprojection loss are adopted for the better handling of occlusions. Mono2 has achieved a huge breakthrough, providing a standard framework for subsequent monocular deep self-supervised tasks. A large number of subsequent tasks are launched with Mono2 as the benchmark. Based on Mono2, Watson ~\textit{et al.}~\cite{watson2021temporal} combine the strengths of monocular and multi-view depth estimation at test time. They introduce efficient losses to improve the accuracy of the moving objects. Guizilini~\cite{guizilini20203d} propose a new convolutional network architecture called PackNet, which is available for high-resolution, and introduce a novel loss that can optionally leverage the camera’s velocity.

\subsection{Multi-Camera Depth Estimation}
Multi-camera depth prediction includes stereo matching and multi view stereo. Stereo matching has been an active field of research for decades~\cite{scharstein2002taxonomy}. Traditional methods utilize handcrafted schemes to find local correspondences~\cite{hosni2012fast}. Recently, based on the deep learning, Goard ~\textit{et al.}~\cite{godard2017unsupervised} take networks to estimate depth with left-right consistency. ~\cite{tankovich2021hitnet} propose a fast multi-resolution initialization step that computes high resolution matches using learned features. However stereo pairs need two cameras being coplanar after rectifying. Multi view stereo reconstruct 3D information of the scene from pictures of different angles. Gu ~\textit{et al.}~\cite{gu2020cascade} apply the cascade cost volume to the MVS-Net, and achieves the best performance on the benchmark. Yang ~\textit{et al.}~\cite{yang2020cost} propose a cost-volume based, compact, and computational efficient depth inference network for MVS. Their framework can handle high resolution images with less memory requirement, and achieve a better accuracy. Khot ~\textit{et al.}~\cite{khot2019learning} propose a self-supervised multi-view stereo architecture  using only images from novel views as supervisory signal. Although these approaches achieve very empirical results, these schemes require large areas of overlap between images. The reduction in the overlap range can have a very serious impact on these methods. Guizilini ~\textit{et al.}~\cite{9712255} innovatively propose full surround monocular depth estimation from multiple cameras in autonomous driving scenarios. By studying spatial-temporal contexts, pose consistency constraints and the effects of overlapping, they open a new direction for multi-camera depth estimation.

%\subsection{Omnidirectional Depth Estimation}

%\subsection{Multi-Camera Datasets}
%DDAD~\cite{guizilini20203d} and NuScenes~\cite{caesar2020nuscenes} are two full surround datasets of autonomous driving. They contains six high-resolution calibrated cameras, which together produce a 360 degree coverage around the vehicle. The datasets contain monocular videos of each camera and accurate ground-truth depth (across a full 360 degree field of view) generated from high-density LiDAR. The six cameras are 2.4MP (1936 $\times$ 1216) in DDAD and 1.4MP (1600 $\times$ 900) in NuScenes, and global-shutter. As shown in Fig.~\ref{fig:ddad overlap}, the six cameras are oriented at about 60 degrees intervals and FOV is about 90 degrees, which creates a small amount of overlap between adjacent cameras. The datasets provide the intrinsics of each camera and the pose parameters between cameras.

\begin{figure}[t!]
\centering
\subfloat[Left view camera.]{
\includegraphics[width=0.32\linewidth]{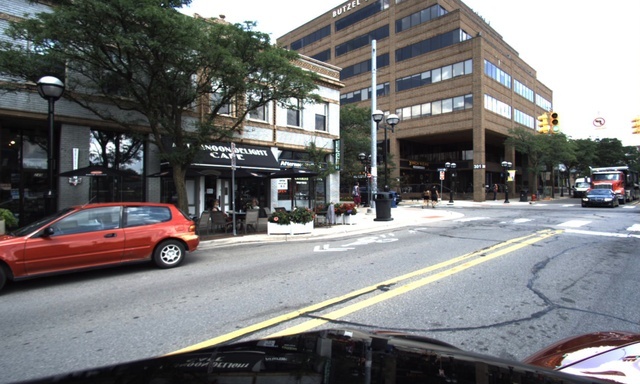}
\label{fig:camera l}
}
\subfloat[Front view camera.]{
\includegraphics[width=0.32\linewidth]{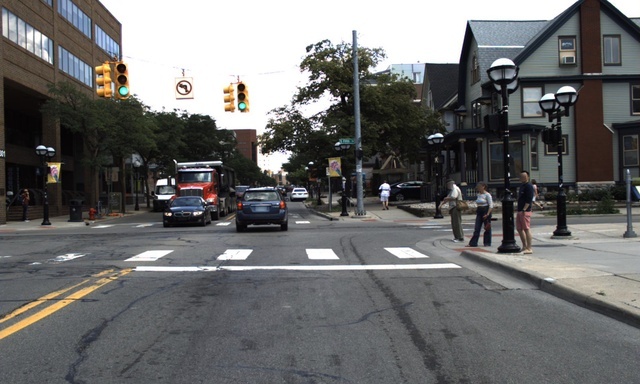}
\label{fig:camera f}
}
\subfloat[Front view synthesized with left view.]{
\includegraphics[width=0.32\linewidth]{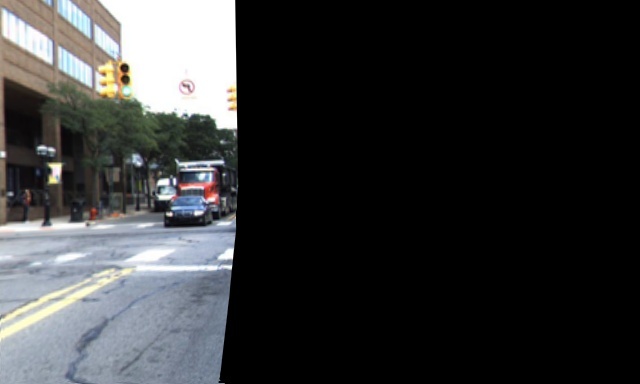}
\label{fig:synthesized f}
}

\caption{Synthesized images across cameras.}
\label{fig:Cross_Cam_Photometric}
\end{figure}

\section{Methodology}
To motivate the design of multi-camera depth estimation approach under the circumstance of small camera overlapping, we first describe the strength and weakness of existing approaches in Sec.\;3.1. We next explain our approach by extending the single-camera monocular self-supervised depth estimation to multi-camera setting in Sec.\;3.2. Our approach can effectively estimate depth maps while maintain structure consistency between different cameras.

%We first describe the standard method to single camera monocular self-supervised depth and ego-motion learning. Then we extend single camera method to multi-camera and introduce two important innovations about multi-camera self-supervised depth estimation approach.

\begin{figure*}[t!]
%\begin{document}
\centering
%\includegraphics[width=1.0\linewidth]{figures/net_architecture/architecture.png}
%\includepdf[pages=1]{figures/net_architecture/architecture.pdf}
\includegraphics[width=18.0cm]{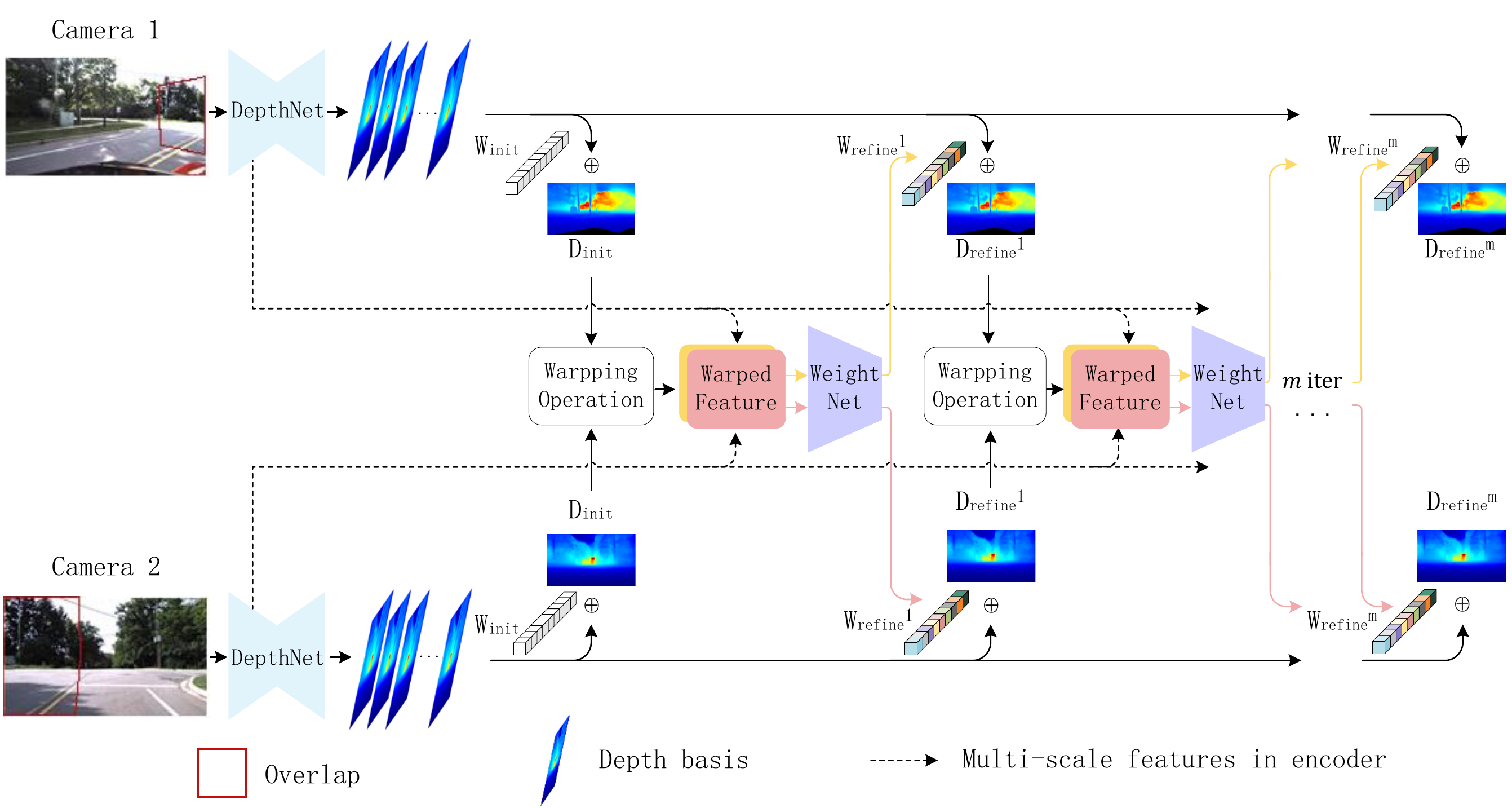}
%\caption{The architecture of the proposed depth basis for consistent structure estimation in multi-camera systems. 
\caption{The architecture of the proposed multi-camera collaborative depth prediction via consistent structure estimation.
Two adjacent cameras estimate the depth maps together. A warped feature contains all the features of one camera and the features of another camera in the overlapping part. The two warped features are respectively inputted into the weight network to calculate the weight of the corresponding camera. $m$ is the refine times. After each refinement, an optimized depth map will be obtained, which is  calculated by the linear combination $\oplus$ of depth basis and weight. $W_{init}$ and $W_{refine^k}$ denote the initial weight and the weight of the k-th refinement.}
\label{fig:net architecture}
\end{figure*}

\subsection{Motivation}

It is challenging to perform multi-camera depth estimation where only small overlapping areas (as low as $10\%$) exist. Recent method FSM~\cite{9712255} proposes a feasible solution based on the standard self-supervised depth monocular estimation, and imposes the constraints on the loss of multi-camera spatio-temporal contexts and multi-camera pose consistency. As shown in Fig.~\ref{fig:Cross_Cam_Photometric}, for two cameras $C^i$ and $C^j$ with overlapping areas, FSM computes the photometric loss between $I^i$ and the synthesized image $I^{j{\rightarrow}i}$ from camera $C^j$ at the same frame. However, such approach suffers from some limitations. By comparing Fig.~\ref{fig:camera f} and Fig.~\ref{fig:synthesized f}, although the structures of the two images in the overlapping area is consistent with each other, there exists significant difference with respect to the brightness level. Besides, some objects existing in one camera are obscured in another camera, due to the large camera placement angle.
Limited by these undesirable artifacts, the synthesized picture cannot be used as an effective supervision signal.%, and photometric loss across cameras reduces the consistency of depth estimation. 
~In addition, FSM only conducts multi-camera collaboration in the training phase while degrades to a single-camera monocular depth estimation approach in the inference phase, which also leads to ineffective exploitation of the information across multiple cameras.

\subsection{Our Approach}

Motivated by the above analysis, we propose an effective scheme to fully employ the cross-camera correlation information in both the training and inference phases.
%\subsubsection{Overall Framework}
The overall framework of the proposed approach is illustrated in Fig. \ref{fig:net architecture}.

\subsubsection{Self-supervised Monocular Depth Estimation}
We first adopt self-supervised monocular depth estimation to predict the depth map from a RGB image without ground-truth, inspired by \cite{godard2019digging}. Given a single input image ${\emph{I}_t}$, we train a baseline network to predict its corresponding depth map ${\emph{D}_t}$. The pose network takes temporally adjacent images as input and estimates relative pose ${T}_{t{\rightarrow}t+n}$ between the target image ${\emph{I}_t}$ and source images ${\emph{I}_{t+n}}$, $n$ $\in$ \{$-1$, $+1$\}. Based on the estimated depth ${\emph{D}_t}$, the relative camera pose ${T}_{t{\rightarrow}t+n}$ and camera intrinsics matrix $K$, we perform view synthesis as the supervisory signal,
\begin{equation}
I_{t+n \rightarrow t}=I_{t+n}\left\langle\operatorname{\emph{proj}}\left(D_{t}, T_{t \rightarrow t+n}, K\right)\right\rangle.
\label{eq: proj}
\end{equation}
Here $\emph{proj}(\cdot)$ are the resulting 2D coordinates of the projected depths ${\emph{D}_t}$ in ${\emph{I}_{t+n}}$, and ${\left\langle \right\rangle}$ is the sampling operator. This view synthesis operation is fully differentiable, enabling gradient back-propagation for end-to-end training. We train the self-supervised monocular depth estimation network by minimizing the per-pixel minimum photometric re-projection error $L_p$ ~\cite{godard2019digging} between the actual target image $\emph{I}_t$ and ${\emph{I}_{t \rightarrow t+n}}$
\begin{equation}
L_{p} =\min _{t+n} pe\left(I_{t}, I_{t+n \rightarrow t}\right),
\label{eq: photometric loss}
\end{equation}
where $pe()$ denotes the photometric error consisting of $L_1$ error and the Structural Similarity (SSIM)~\cite{wang2004image}
\begin{equation}
\begin{split}
pe\left(I_a, I_b\right) = &\frac{\alpha}{2}\left(1-\operatorname{SSIM}\left(I_{a}, I_{b}\right)\right) \\
&+(1-\alpha)\left\|I_{a}-I_{b}\right\|_{1},
\end{split}
\label{eq: photometric rec loss}
\end{equation}
where $\alpha$ = 0.85. 
%\textcolor{red}{As in~\cite{godard2017unsupervised}, we use edge-aware smoothness where $d^*_t = d_t / \overline{d_t}$ is the mean-normalized inverse depth
As in~\cite{godard2017unsupervised}, we use edge-aware smoothness which calculates the mean-normalized inverse depth
from \cite{wang2018learning} to discourage shrinking of the estimated depth.

%\end{document}

\subsubsection{Multi-Camera Collaborative Depth Prediction}
\label{sec:Multi-camera framework}
Multi-camera approaches to self-supervised depth estimation are severely limited by camera placement setting. Stereo depth estimation methods~\cite{badki2020bi3d,kusupati2020normal} need to rectify the images to predict the disparities with a known baseline. The approaches to multi-view stereo~\cite{yang2020cost,gu2020cascade} predict depth maps based on different camera positions to build the cost volume. Although these methods have been proposed to utilize the positional relationship and information between images, they all need a large overlapping areas in the images, such as KITTI~\cite{geiger2012we}, DTU~\cite{aanaes2016large} datasets.

\begin{figure}[t]
\centering
\subfloat[Left view image.]{
\includegraphics[width=0.32\linewidth]{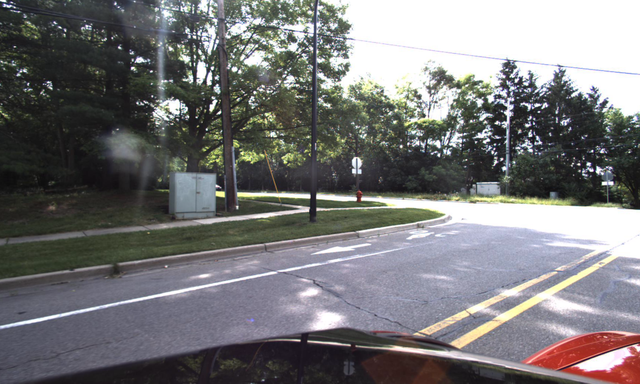}
\label{fig:project rgb l}
}
\subfloat[Front view image.]{
\includegraphics[width=0.32\linewidth]{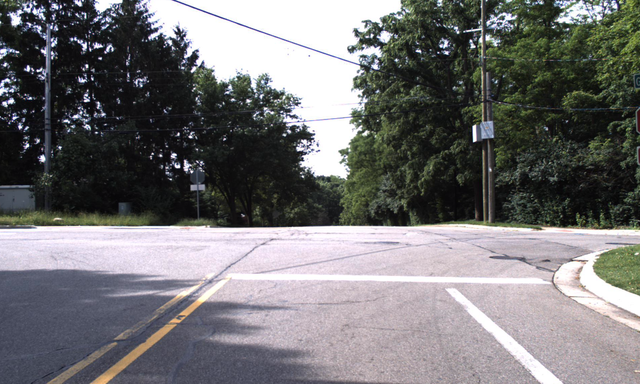}
\label{fig:project rgb m}
}
\subfloat[Right view image.]{
\includegraphics[width=0.32\linewidth]{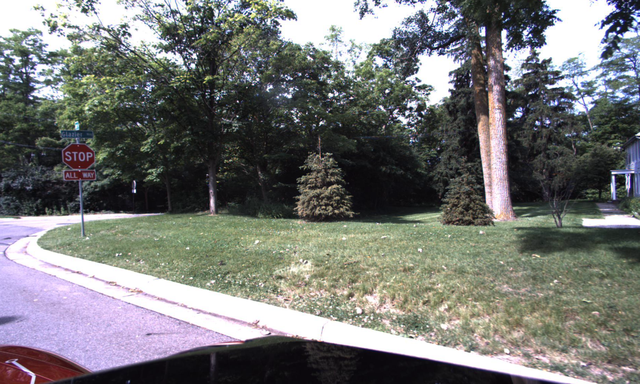}
\label{fig:project rgb r}
}
\\
\subfloat[Left view depth.]{
\includegraphics[width=0.32\linewidth]{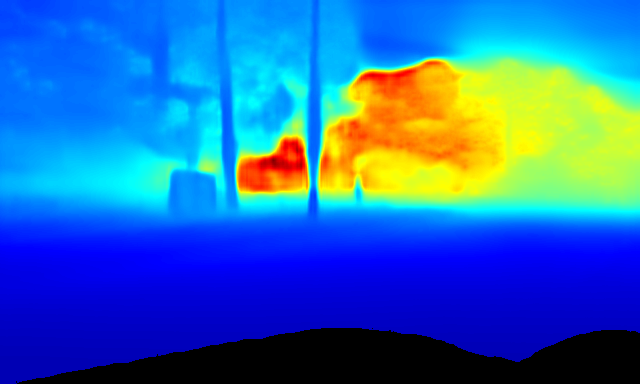}
\label{fig:project depth l}
}
\subfloat[Depth projected from left and right views.]{
\includegraphics[width=0.32\linewidth]{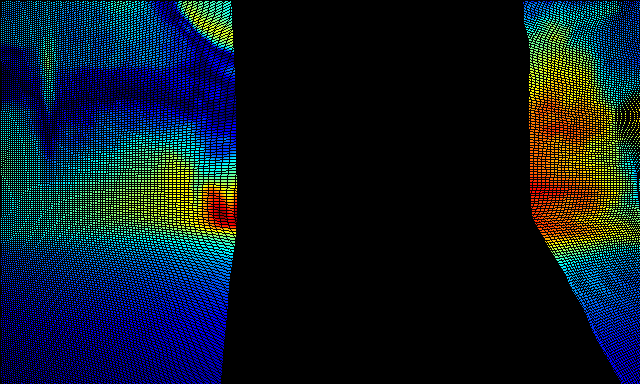}
\label{fig:project depth f}
}
\subfloat[Right view depth.]{
\includegraphics[width=0.32\linewidth]{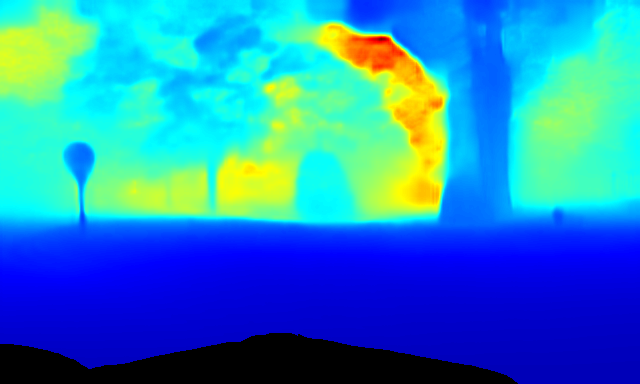}
\label{fig:project depth r}
}
\caption{Depth projection from different views}
\label{fig:depth project}
\end{figure}

%we find that  many scenarios pictures taken from different angles at the same time have problems such as different brightness, occlusion and so on. As shown in Fig.~\ref{fig:Cross_Cam_Photometric}, there is a great difference in brightness between the synthetic image ~\ref{fig:synthesized f} by ~\ref{fig:camera l} and the real image ~\ref{fig:camera f}.
%So when there is a large angle between the cameras, it is not appropriate to use photometric loss between adjacent cameras.
\paragraph{Depth Basis for Consistent Structure Estimation}
We propose the depth basis for consistent structure estimation to solve the problem of self-supervised depth estimation consistency with small overlapping areas in multi-camera systems. It can make more rational use of the overlapping images, and use the information of adjacent cameras to improve the accuracy of overall depth estimation in the inference stage.

Inspired by ~\cite{bloesch2018codeslam}, the dense depth map can be adjusted by a small number of basic states, and we modify the architecture of monocular depth estimation network.  
Specifically, the image $I^{i}_t$ from camera $C^i$ is fed to the modified depth estimation network to generate $n$ depth bases 
$B^{i}\in \mathbb{R}^{n\times H \times W}$
instead of the depth, where $H$ and $W$ are the height and width of the input images. Each basis represents a state of possible distribution of the depth. The estimated depth map is the combination of the bases.
We set $n$ initial weights $W_{init}^{i}$ and calculate the initial depth $D_{init}^{i}$ with $B^i$
\begin{equation}
D^{i}_{init}=B^{i} \oplus W_{init}^{i}=\sum_{j=1}^n w_{init}^{i,j}\cdot B^{i,j},
\label{eq: init depth}
\end{equation}
where $w_{init}^{i,j}$ is the $j$-th element of $W_{init}^{i}$ and $B^{i,j}$ is the $j$-th depth basis of $B^i$. For each camera, the initial weight $w_{init}^{i,j}$ is set to $\frac{1}{n}$. The size of $D_{init}^{i}$ is 1 $\times$ $H$ $\times$ $W$, which is the same as each basis. The $n$ depth bases enjoy more flexibility to express the depth, which provides a place to optimise.
If we fix the initial weights $W_{init}^{i}$, there is no difference between generating depth directly and using bases to linearly compute depth as an optimization problem. Another advantage is that we only need to adjust the weights of bases to refine the combination instead of recalculating the depth map pixel by pixel. This greatly reduces the number of parameters and computational complexity.
%and we can refine the depth by controlling the weight.
When more than one camera are available, the weight can be further optimized by exploiting the overlapping views between the target camera and the adjacent cameras.

%depth map cannot be further optimized, updating new weight needs more information.  When adjacent cameras shooting from another perspective, we obtain additional features in the overlapping part of the images, which making the refinement operation possible.
%overlapping regions provided by adjacent cameras will provide additional information, making the refinement operation possible.

\paragraph{Weight Network.}
We assume that the cameras are rigidly connected in the multi-camera setting, and obtain the extrinsics matrix ${\mathbf{E}}^{i \to j}$ including ${R}^{i\to j}$ and ${t}^{i\to j}$
%${\mathbf{E}}^{i \to j} = \begin{psmallmatrix}\mathbf{{R}^{i\to j}} & \mathbf{{t}^{i\to j}}\\ \mathbf{0} & \mathbf{1}\end{psmallmatrix} \in \text{SE(3)}$ 
between each two cameras from the datasets. Combining intrinsics $K$ of the standard pinhole model~\cite{page2005multiple} and the estimated depth $D$, we obtain the pixel-warping operation between the camera  $C^i$ and $C^j$:
\begin{equation}
\hat{\mathbf{p}}^i =
\pi_j \big(\mathbf{R}^{i \rightarrow j} \phi_i (\mathbf{p}^i, D^i,{K}^{i}) + \mathbf{t}^{i \rightarrow j},{K}^{j}\big),
\label{eq:warp_spatial}
\end{equation}
where $\phi(\mathbf{p}, D, K) = \mathbf{P}$ is the unprojection of a pixel in homogeneous coordinates $\mathbf{p}$ to a 3D point $\mathbf{P}$ for a given estimated depth $D$. $\pi(\mathbf{P},{K}) = \mathbf{p}$ denotes the projection of a 3D point back onto the image plane.
%Both operations require the camera parameters, which for the standard pinhole model  \cite{hartley2003multiple} is defined by the $3 \times 3$ intrinsics matrix $\mathbf{K}$.

To exploit the useful information of the overlapping parts, we warp feature maps of another view obtained by the encoder in the depth network. We first obtain features at three scales in the encoder, $H/{2^{i}} \times W/{2^{i}}$, ($i=1, 2, 3$, $H \times W$ is the size of the input image). We use bilinear interpolation to align the multi-scale features to the same size $H/2 \times W/2$ and concatenate them along channels. Based on the pixel-warping operation between cameras \eqref{eq:warp_spatial}, we warp the multi-scale features $F^{j}$ from $C^{j}$ to $C^{i}$.
The non-overlapping area of the image is filled with $0$.
We concatenate the warped $F^{j}$ with $F^i$, leading to the feature $\hat{F}^i$.
In the view of $C^i$, $\hat{F}^i$ has not only all the features of $I^i$, but also the multi-scale features of the overlapping part from $I^j$. 

\begin{figure}[t]
  \begin{center}
    \small

    \includegraphics[width=0.9\linewidth]{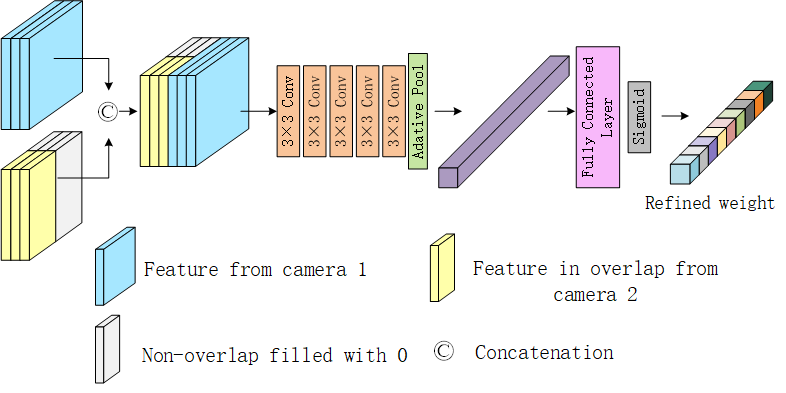}
  \end{center}
  \caption{The architectures of the weight network}
  \label{fig:WeightNet}
  %\vspace{2mm}\hrule
\end{figure}

With the more informative feature $\hat{F}^i$, we propose to refine the combination of the depth basis in order to generate a higher quality depth map. 
We design a \emph{weight network} in order to refine the weights for the combination of depth bases $B^i$.
%It is very complicated for features to guide the adjustment of the $B$, so we design the \emph{weight network} to execute this function. 
The input of the network is $\hat{F}^i$ and the output is the refined weight $W^i_{refine}$. Thanks to the new features provided by adjacent cameras, the weight network can achieve a more accurate solution to the depth map estimation of the target view. We show the specific structure of the weight network in Fig.~\ref{fig:WeightNet}. 
%The first half $Weight\ Network$ is composed of multiple layers of convolution, batch normalization and activation function. After adaptive average pooling and three fully connected layers, we can get the refined weight $W_{refine}^i$.
With $W_{refine}^i$ and $B^i$, we can generate the refined depth map $D_{refine}^i$ with \eqref{eq: init depth}.

%The same operation as generating the initial depth map with the depth basis \eqref{eq: init depth}, the new weights result in a refined depth map. Because $D_{refine}^i$ is generated from multi-camera, it has richer information than monocular prediction, and the $W_{refine}$ will make the combination of $B$ more reasonable.

\paragraph{Iterative Refinement Strategy}
After one refinement, we can get a more accurate depth map $D^i_{refine}$. According to \eqref{eq:warp_spatial}, the quality of depth map impacts the pixel-warping operation, and the refined depth map can reduce the projection errors.
We thus propose a iterative refinement strategy to use the refined depth map in the current step as the input of the next step, and repeat this process for $m$ times. The images from different cameras shares the same weight network in each step, but the networks of different steps are independent from each other.
We analyze the effect of the number $m$ of iterations in the experiments.

%Referring to the first $D_{refine}$ refinement process, each time the optimized depth map can be refined again through the warpping operation and the weight network, and repeat this process iteratively. On the implementation details,  images from different cameras use a shared weight network in each  process of optimization, and the networks are independent between iterations.
%In the experiment of the next chapter, we will analyze in detail the effect of the number of iterations on the experimental results.

\subsubsection{Loss Function}
The overlap taken by the multi-camera predicts the depth map of each image.
%The overlapping area is estimated multiple times by multiple cameras, and the projected depths predicted by different cameras must have errors in the real scene.
The consistency between multiple depth maps has a great impact on downstream tasks, such as autonomous driving. Low depth consistency can lead to semantic error, path planning deviation and so on. To achieve more consistent depth estimation for overlapping areas under different cameras, Fan ~\textit{et al.} come up with \emph{Chamfer Loss}~\cite{fan2017point}, which can backproject the depth map to the world coordinate system. It improves consistency by minimizing the distance between two point clouds in overlapping areas. But the point cloud generated by the depth map backprojection is very dense. The resolution of the image is  $1936\times  1216$ in DDAD dataset~\cite{guizilini20203d}, and there are more than 2 million points in one point cloud generated by an image. Although \emph{chamfer distance} can be optimized by $KD\ Tree$, computation between point clouds consumes a lot of computing power. In our experiment, a standard depth network iteration spends 0.5 about seconds, but it spends over 10 seconds with \emph{chamfer loss}.

\paragraph{Depth Consistency Loss} We design a depth consistency loss $L_{con}$ to reduce the error between the estimated depth maps from different cameras in the overlap. Since the camera planes of different cameras are not coplanar, the depth maps estimated from different cameras cannot be directly compared. We 
%thus
project the depth values onto the same camera plane by \eqref{eq:warp_spatial}. Instead of grid sampling and interpolation operation, we use the simple coordinate transformation. We can project $D^j$ from $C^j$ to $C^i$ and generate a depth map $\hat{D}^i$ with only overlapping areas predicted by the $C^j$. As shown in Fig.~\ref{fig:depth project}, the depth maps in Fig.~\ref{fig:project depth l} and ~\ref{fig:project depth r} are estimated by the left view in Fig.~\ref{fig:project rgb l} and right view in Fig.~\ref{fig:project rgb r} respectively, and project them to front view in Fig.~\ref{fig:project depth f}.  In the process of projecting the point cloud to the image plane, multiple points may be projected to the same pixel. Based on pixel order, the points projected later overwrite the points projected earlier, and pixels that are not projected to are filled with $0$.

Under the same coordinate system, we can directly calculate the difference in depth for the overlapping areas. We $L_1$ loss to constraint between $D^i$ and $\hat{D}^i$. We denote depth consistency loss:
\begin{equation}
L_{con} = L_1 (D^i, \hat{D}^i).
\label{eq:Con loss}
\end{equation}

\paragraph{Full Loss.}
Just like the basic self-supervised monocular depth estimation method~\cite{godard2019digging}, we need to predict ego-motion in the image sequence to construct supervision according to \eqref{eq: photometric loss}. The pose network takes each temporally images \{$I_t^{i}$,$I_{t-1}^{i}$,$I_{t+1}^{i}$\} from the camera $C^i$ to estimate the pose $T^{i}_{t{\rightarrow}t+n}$. The depth estimation network is shared between cameras.
%The pose network takes each temporally adjacent images \{$I_t^{i}$,$I_{t-1}^{i}$,$I_{t+1}^{i}$\} from different cameras $C_i$ to estimate the pose $T^{i}_{t{\rightarrow}t+n}$, which is just like the basic monocular self-supervised depth estimation method.
During multi-camera self-supervised training, the full loss is comprised of the photometric loss $L_p$ and depth consistency loss $L_{con}$. Because the depth maps resulting from the MCDP include initial depth map and multiple refined depth maps, the full loss takes the form:
\begin{equation}
L = L_p^{init} + \lambda L_{con}^{init} +  \sum_{k=1}^{\mathit{m}}(L_p^{k} + \lambda L_{con}^{k}),
\label{eq:Loss}
\end{equation}
where $\lambda$ is a hyper-parameter to balance the $L_{p}$ and $L_{con}$, and $m$ is the number of refinement steps. We assume that each refinement step contributes equally in \eqref{eq:Loss}.

%define the same weight between the initial result and the result after each refinement.

%\textbf{Depth Consistency Evaluation.}
%How to evaluate the consistency of depth estimation in overlapping regions of multiple cameras is an important issue. 
%Based on the proposed depth consistency loss, we design a new depth consistency metric \textit{Dep\ Con} to evaluate depth consistency in the image plane:
%\begin{equation}
%Dep\ Con =\frac{ |D^i - \hat{D}^i|}{D_{gt}},
%\label{eq:Dep Con}
%\end{equation}
%where $D_{gt}$ is the ground-truth depth map. By \eqref{eq:warp_spatial}, we project the  depth maps of overlaps to the same camera plane, calculating the ratio of the absolute errors of $D^i$ and $\hat{D}^i$ to the ground-truth. The role of the ratio is to balance the effect of the prediction distance on the error. Instead of calculating distances between point clouds, we calculate the distance between two depth maps, which can significantly reduce computational complexity. We use this metric to evaluate our experimental results in the experiments.

\section{Experiments}
\subsection{Datasets}
Traditionally, the methods of monocular self-supervised depth estimation are often verified on the KITTI dataset~\cite{geiger2012we}. There are only rectified stereo pairs from forward-facing cameras in the KITTI. In recent years, some multi-camera omnidirectional autonomous driving datasets have been open sourced. For the setting of multiple cameras and a small amount of overlapping area, we choose the two datasets to validate our scheme.

\paragraph{DDAD \cite{guizilini20203d}} The Dense Depth for Automated Driving (DDAD) is an urban driving dataset captured with six synchronized cameras with relatively small overlap. It contains highly accurate dense ground-truth depth maps for evaluation and max depth range is up to 250 meters. It has a total of 12650 training samples (63250 images) and 3950 validation samples (15800 images). In the training set, we do not use ground-truth depth maps. The resolution of the images is 1936 $\times$ 1216. Following the procedure outlined in~\cite{9712255}, input images are downsampled to the 640 $\times$ 384 resolution. For evaluation, we use bilinear interpolation to upsample the image resolution to the original size.

\paragraph{NuScenes~\cite{caesar2020nuscenes}}
The NuScenes dataset is an urban driving dataset that contains images from a synchronized six-camera array. It comprises of 1000 scenes with a total of 1.4 milllion images. It is a popular benchmark for 2D and 3D object detection, as well as semantic and instance segmentation. This dataset is challenging for self-supervised depth estimation task because of the relatively low resolution of the images, very small overlap between the cameras, high diversity of weather conditions and time of day, and unstructured environments. The size of raw images is 1600 $\times$ 900, which are downsampled to 768 $\times$ 448. The images are captured at 30Hz, and the samples of the dataset are annotated at 2Hz as key frames. Due to the large time interval of key frames, deep networks cannot be trained in a self-supervised manner. We use the sequences from the raw dataset as the supervisory signal, which are not annotated.
%We don't need to flip the image left and right to take the average to improve the

\begin{figure}[t]
\centering
%\subfloat[Input RGB image.]{
\subfloat[Left front view.]{
\includegraphics[width=0.4\linewidth]{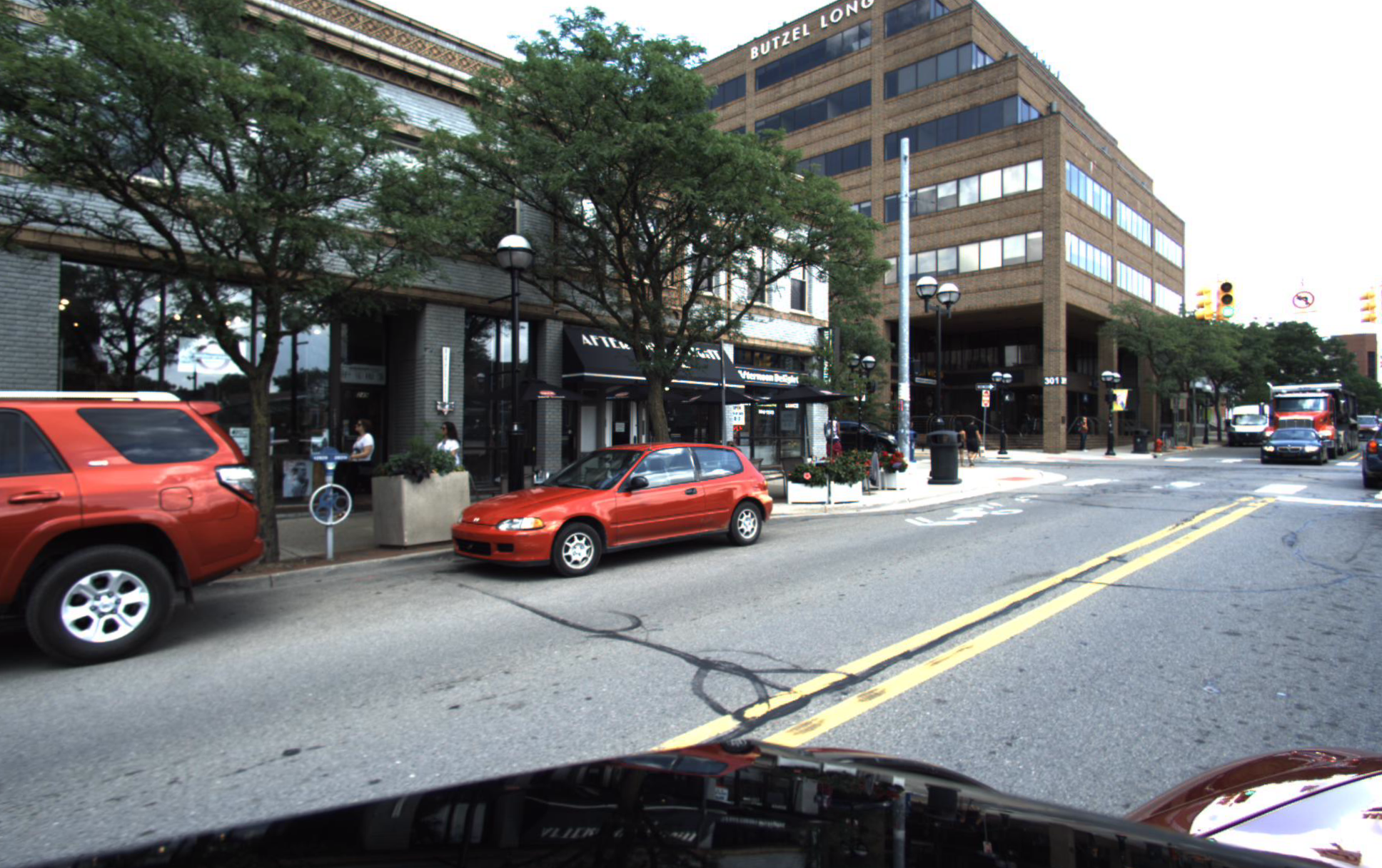}
\label{fig:mask_rgb}
}
\subfloat[Mask of left front view.]{
\includegraphics[width=0.4\linewidth]{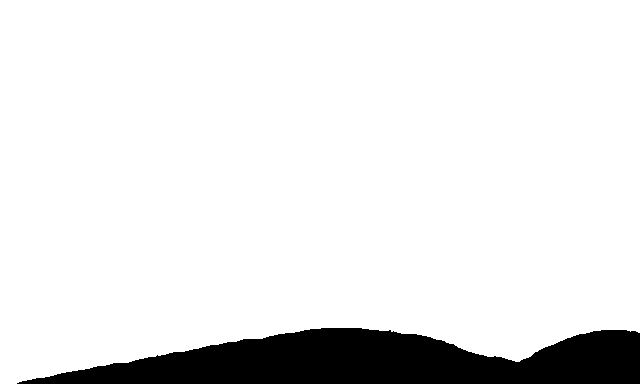}
}
\\
\subfloat[Left back view.]{
\includegraphics[width=0.4\linewidth]{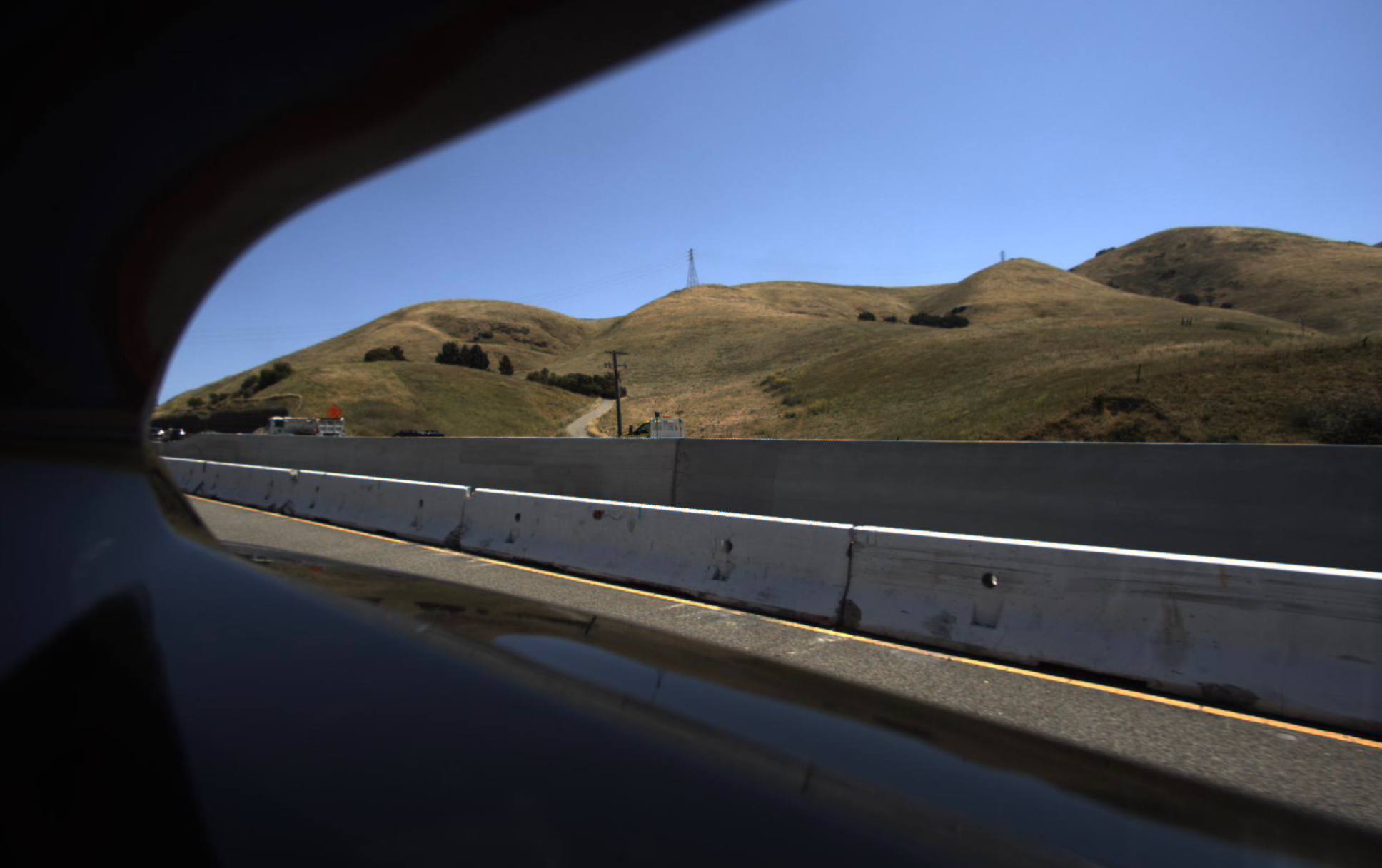}
}
\subfloat[Mask of left back view.]{
\includegraphics[width=0.4\linewidth]{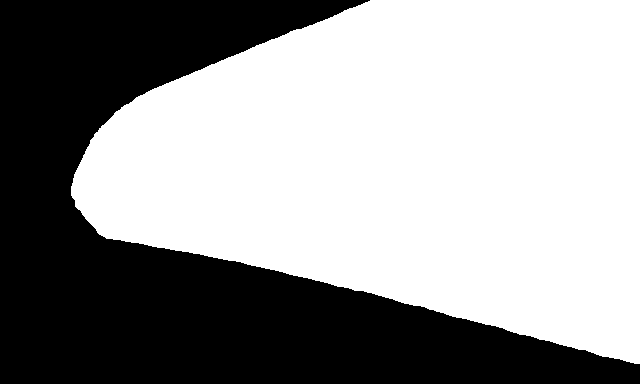} %width=0.44
}
\caption{Self-Occlusions Mask.}
\label{fig:mask}
\end{figure}

\subsection{Multi-Camera Depth Evaluation Metrics}

We take the \emph{median-scaling}~\cite{zhou2017unsupervised} approach to evaluate the results, which is commonly used in self-supervised monodepth estimation at test time. 
%\textbf{Depth Consistency Evaluation.}
%How to evaluate the consistency of depth estimation in overlapping regions of multiple cameras is an important issue. 
Based on the proposed depth consistency loss, we design a new depth consistency metric \textit{Dep\ Con} to evaluate depth consistency in the image plane:
\begin{equation}
Dep\ Con =\frac{ |D^i - \hat{D}^i|}{D_{gt}},
\label{eq:Dep Con}
\end{equation}
where $D_{gt}$ is the ground-truth depth map. We project the depth maps of overlaps to the same camera plane by \eqref{eq:warp_spatial}, and calculate the ratio of the absolute errors of $D^i$ and $\hat{D}^i$ to the ground-truth. The role of the ratio is to balance the effect of the prediction distance on the error. Instead of calculating distances between point clouds, we calculate the distance between two depth maps, which can significantly reduce computational complexity.

In addition to the proposed \emph{Dep Con} metric, we use the standard four metrics used in the prior work~\cite{godard2019digging}: average relative error (Abs Rel), squared relative difference (Sq Rel), root mean squared error (RMS), threshold accuracy ($\delta_{1.25}$).

\iffalse
\begin{itemize}
\item Average relative error (Abs Rel): $\frac{1}{n}\sum_p^n \frac{\lvert y_p-\hat{y}_p \rvert}{y}$;
\item Aquared relative difference (Sq Rel): $\frac{1}{n}\sum_p^n \frac{\|y_p-\hat{y}_p \|^2}{y}$;
\item Root mean squared error (RMS): $\sqrt{\frac{1}{n}\sum_p^n (y_p-\hat{y}_p)^2)}$;
\item Threshold accuracy ($\delta_i$): percentage of $y_p$ s.t.
$\text{max}(\frac{y_p}{\hat{y}_p},\frac{\hat{y}_p}{y_p}) = \delta < thr$ for $thr=1.25$.

\end{itemize}
\fi

\begin{figure*}[th!]
\vspace{-2mm}
\centering
\subfloat{
\includegraphics[width=0.16\linewidth,height=1.5cm]{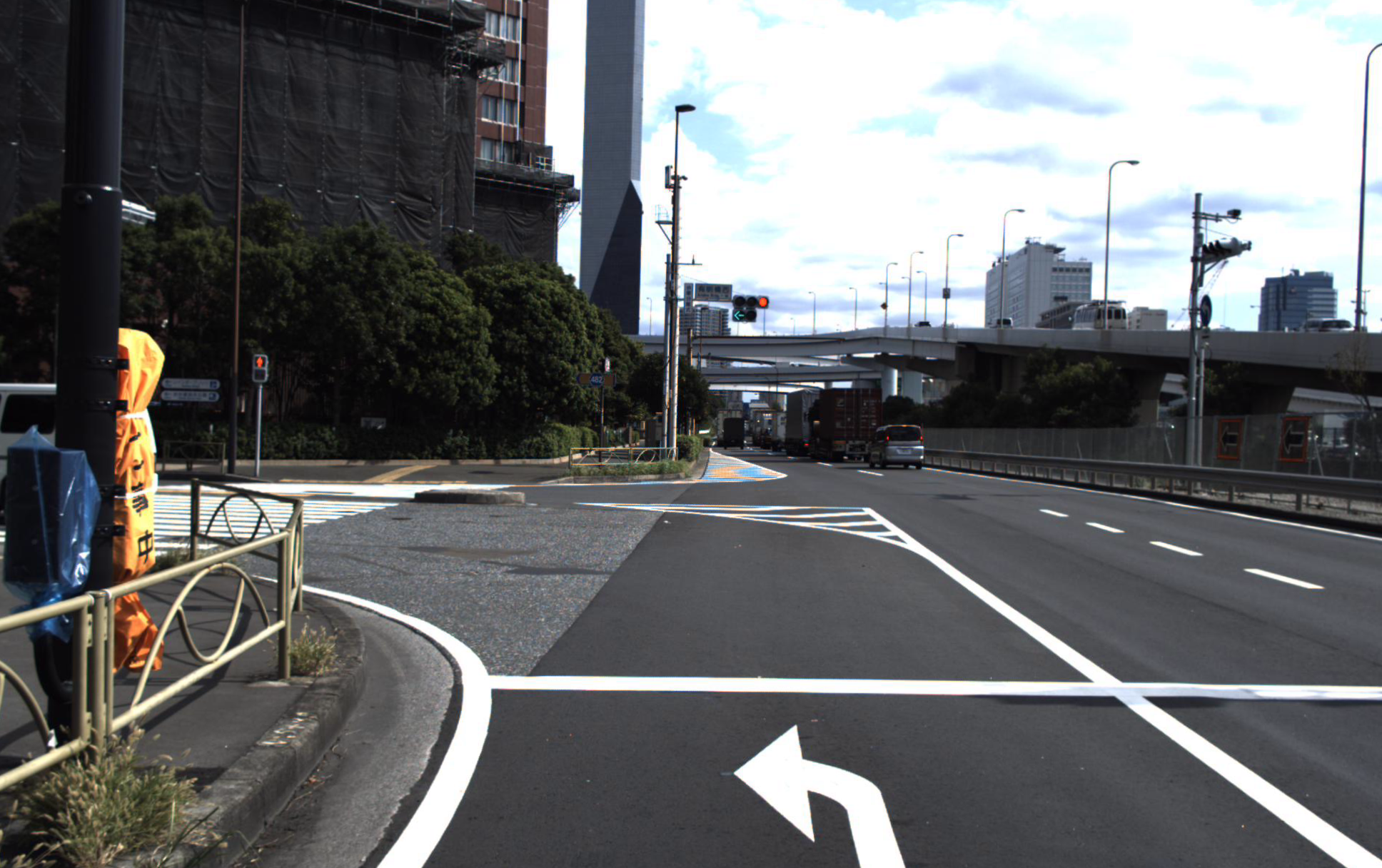}
\includegraphics[width=0.16\linewidth,height=1.5cm]{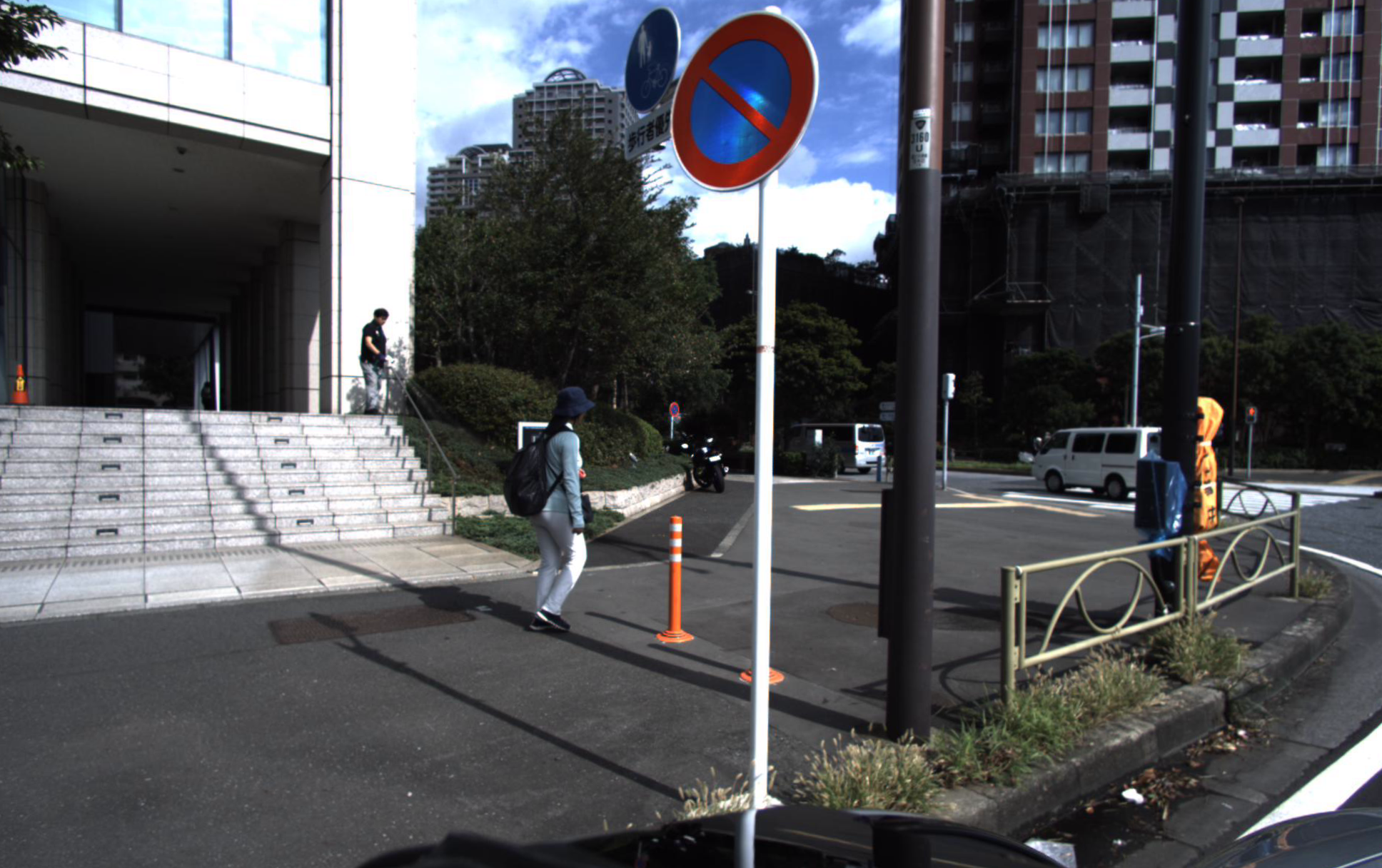}
\includegraphics[width=0.16\linewidth,height=1.5cm]{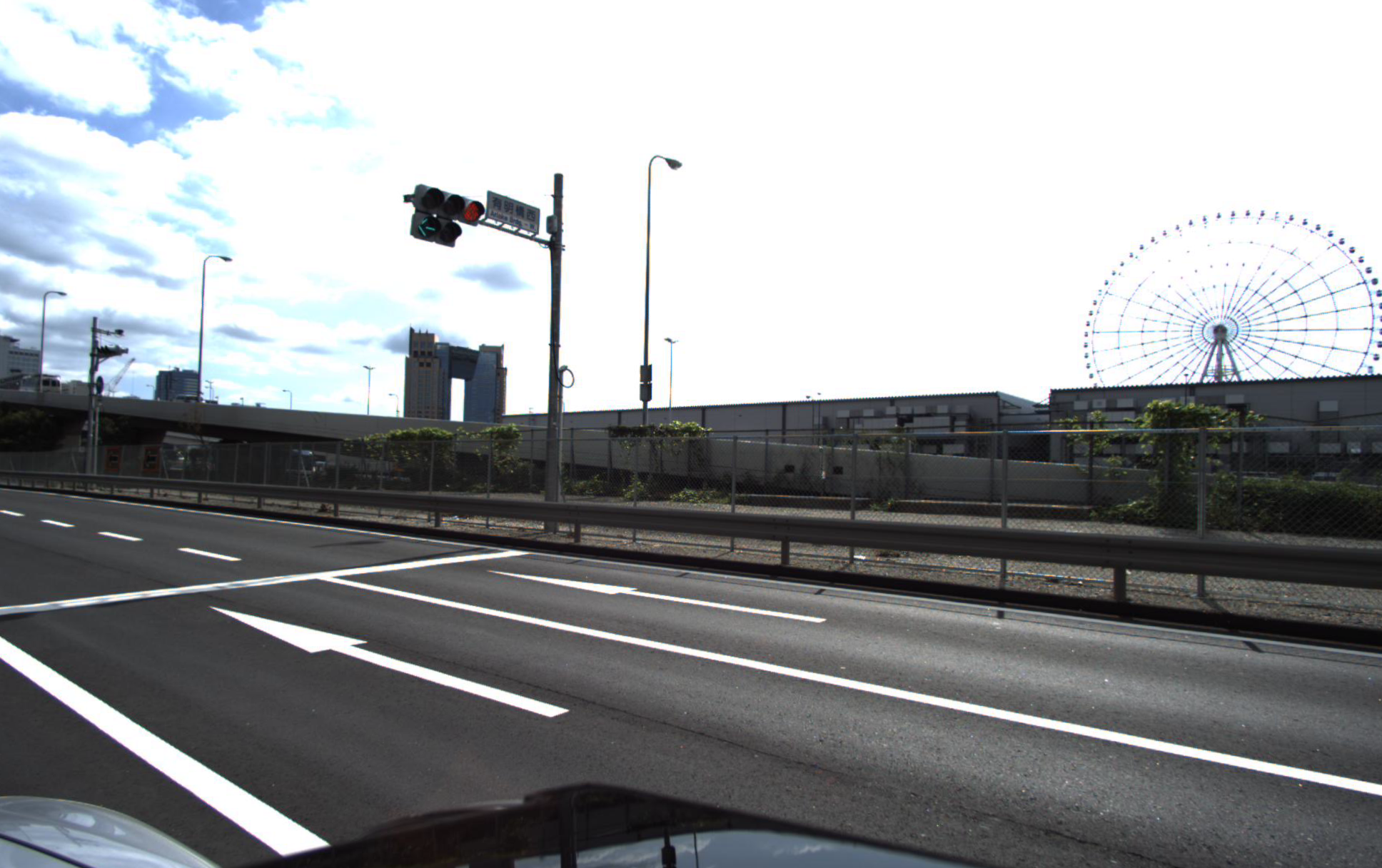}
\includegraphics[width=0.16\linewidth,height=1.5cm]{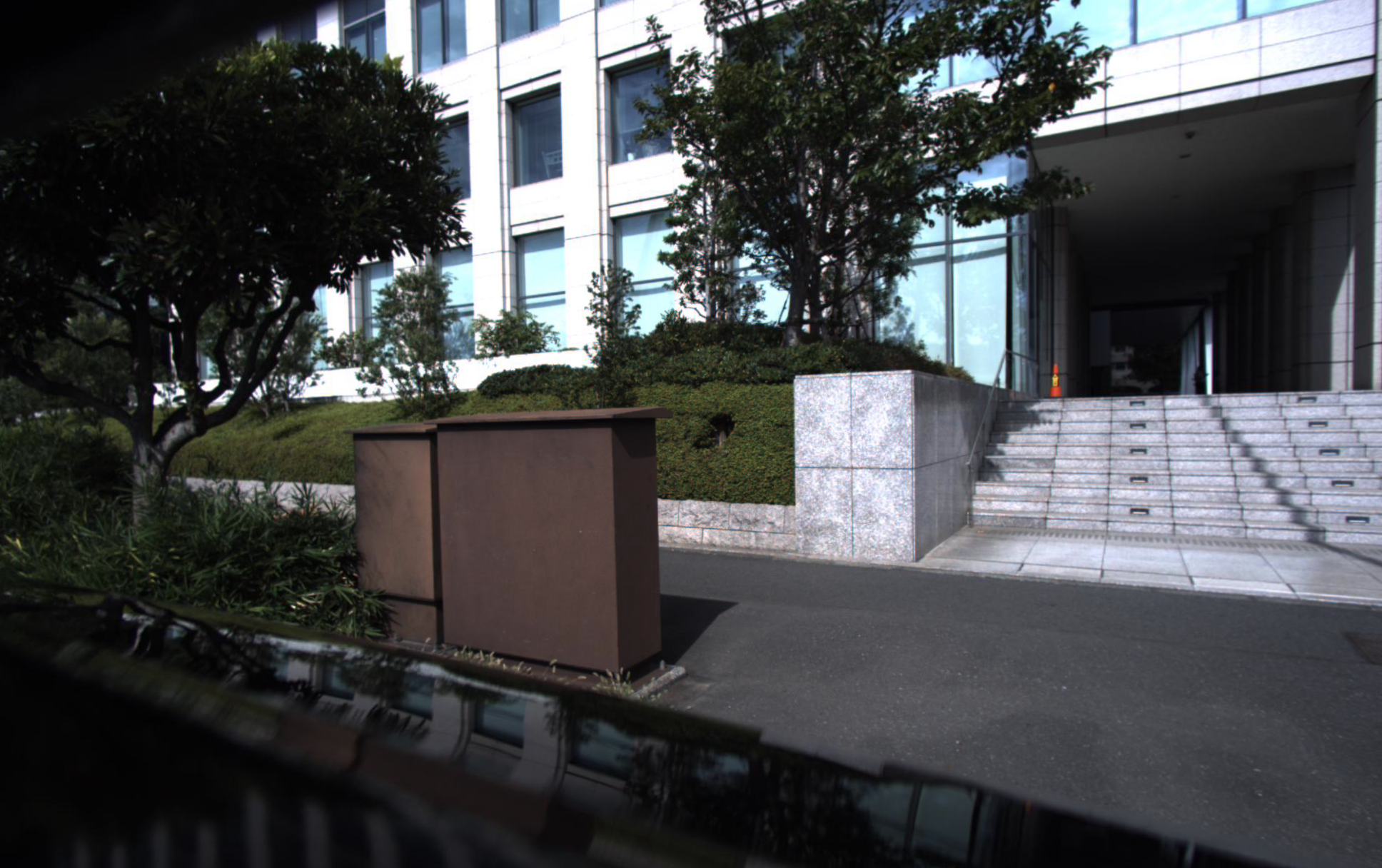}
\includegraphics[width=0.16\linewidth,height=1.5cm]{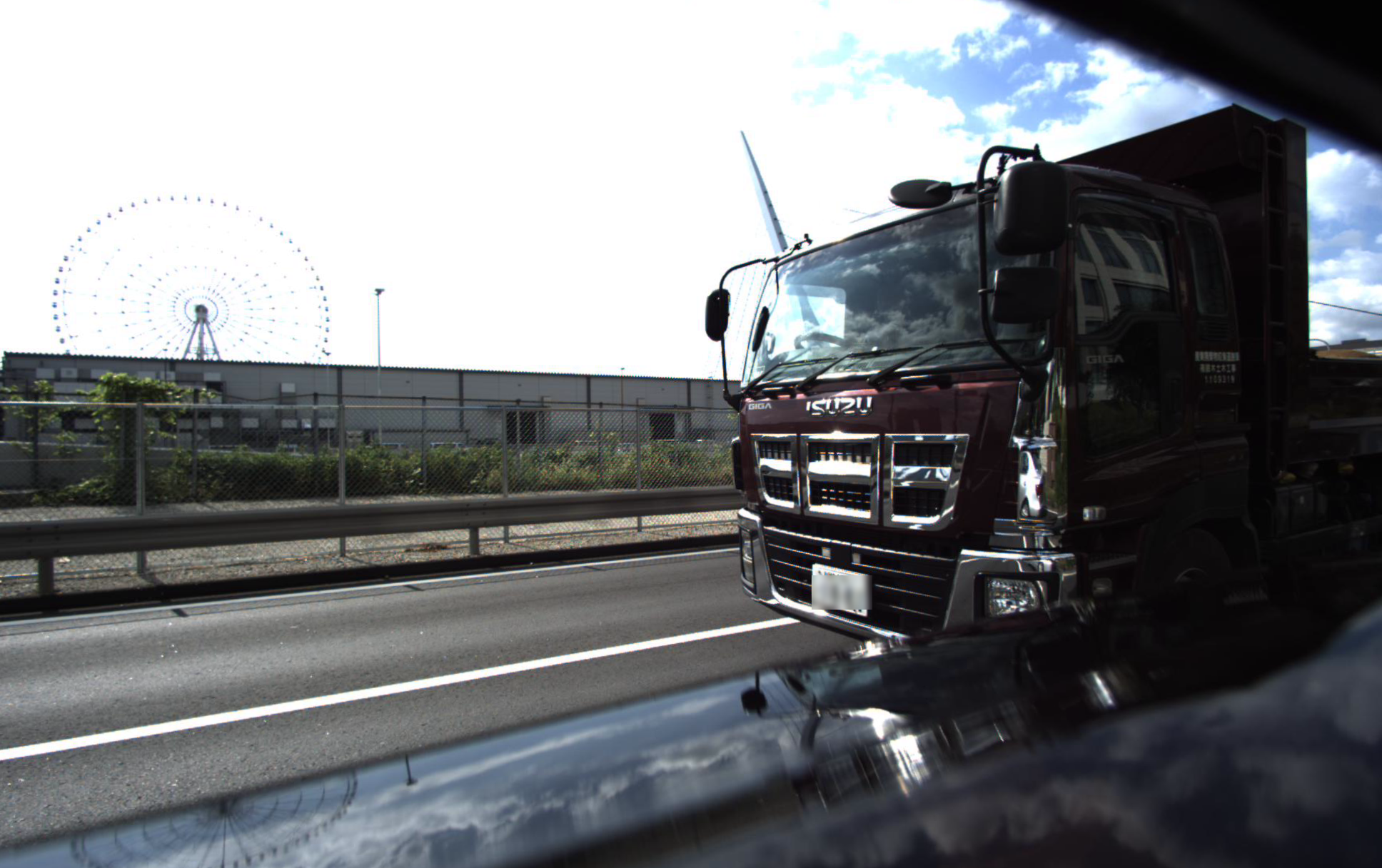}
\includegraphics[width=0.16\linewidth,height=1.5cm]{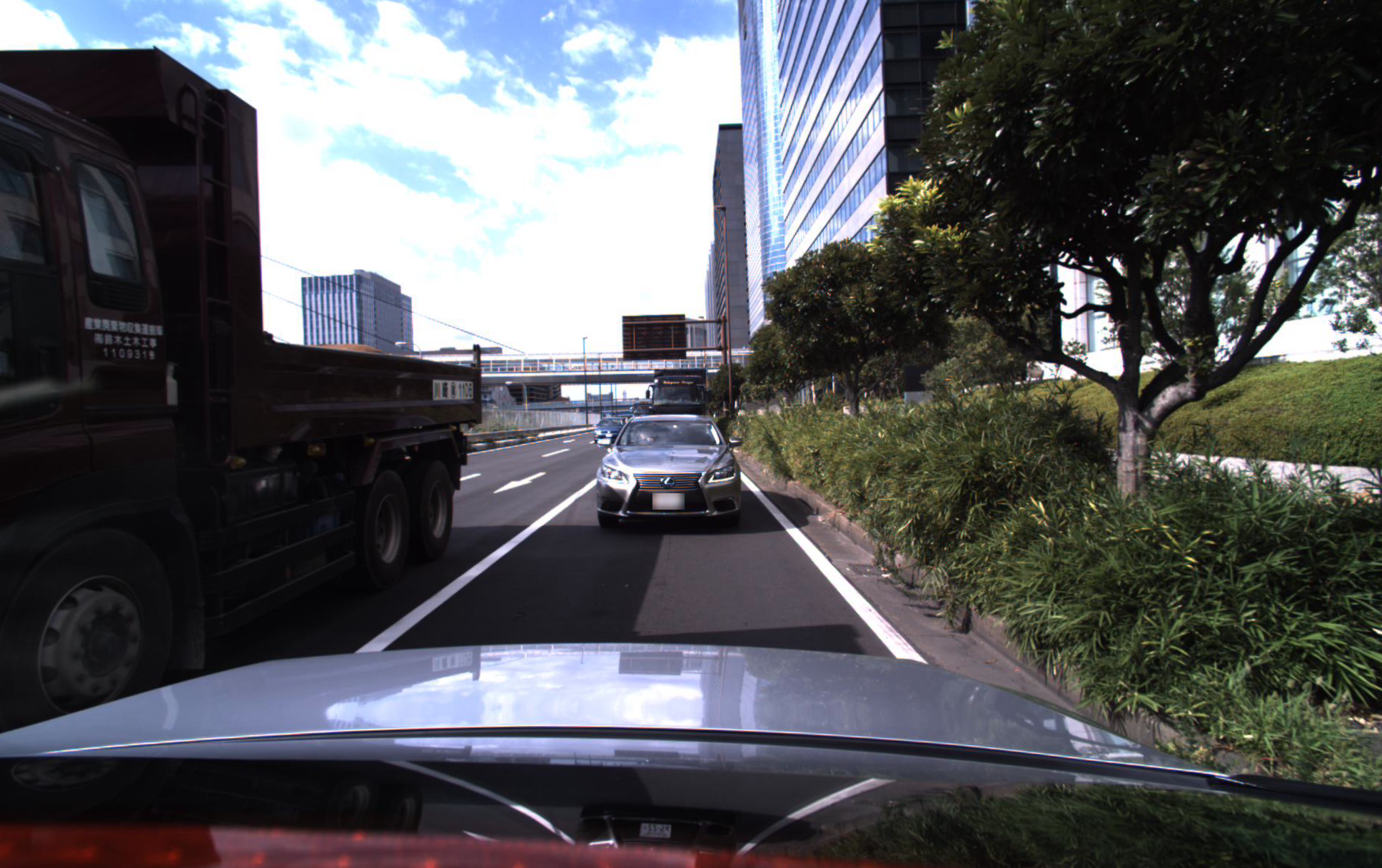}
}
\vspace{-4mm}
\\
\subfloat{
\includegraphics[width=0.16\linewidth,height=1.5cm,height=1.5cm]{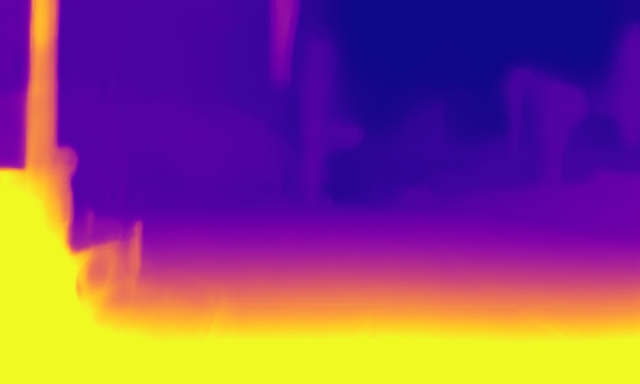}
\includegraphics[width=0.16\linewidth,height=1.5cm,height=1.5cm]{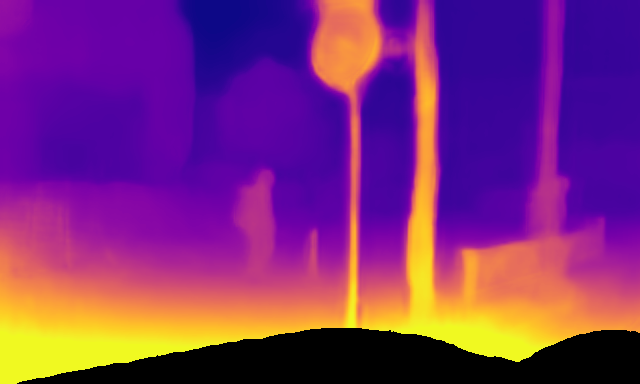}
\includegraphics[width=0.16\linewidth,height=1.5cm,height=1.5cm]{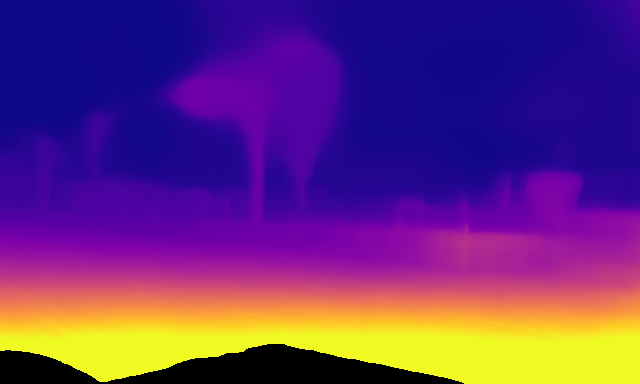}
\includegraphics[width=0.16\linewidth,height=1.5cm,height=1.5cm]{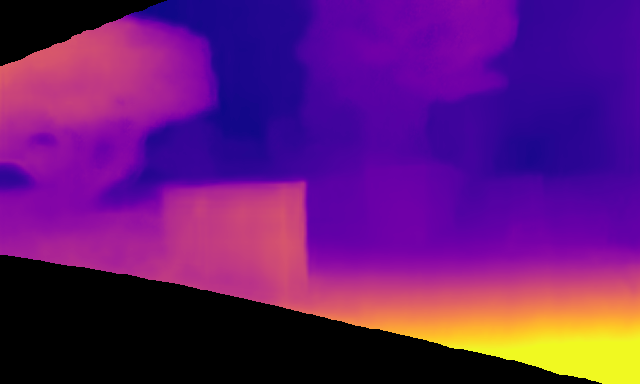}
\includegraphics[width=0.16\linewidth,height=1.5cm,height=1.5cm]{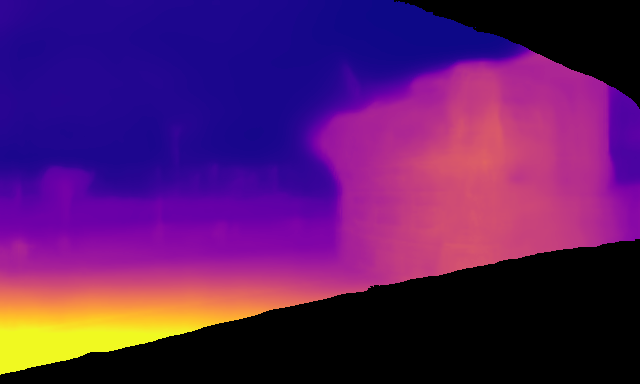}
\includegraphics[width=0.16\linewidth,height=1.5cm,height=1.5cm]{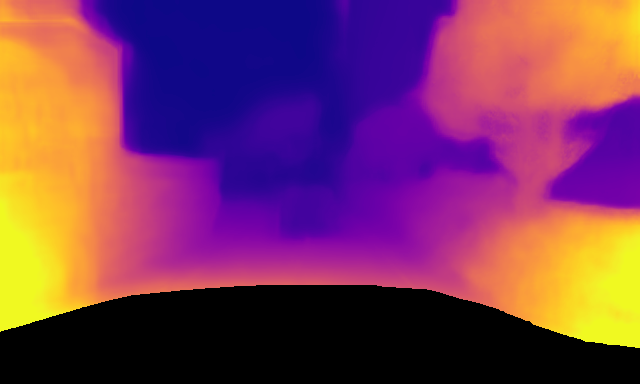}
}
\vspace{-4mm}
\\
\subfloat{
\includegraphics[width=0.16\linewidth,height=1.5cm,height=1.5cm]{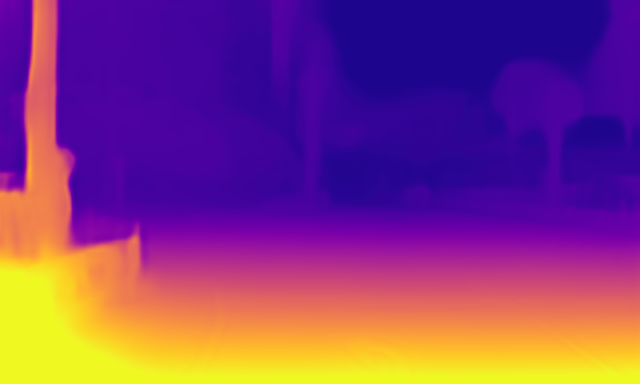}
\includegraphics[width=0.16\linewidth,height=1.5cm,height=1.5cm]{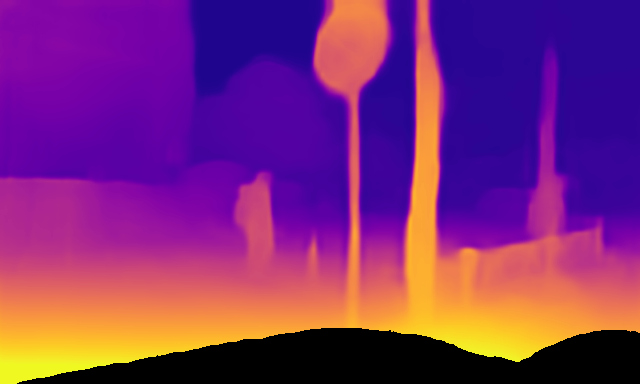}
\includegraphics[width=0.16\linewidth,height=1.5cm,height=1.5cm]{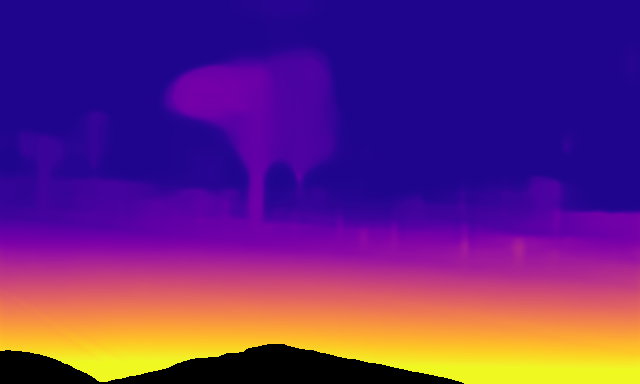}
\includegraphics[width=0.16\linewidth,height=1.5cm,height=1.5cm]{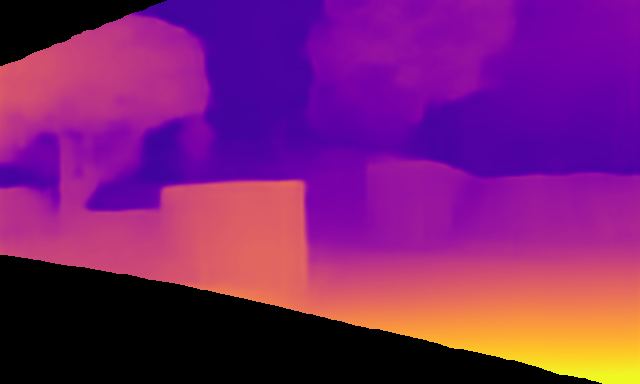}
\includegraphics[width=0.16\linewidth,height=1.5cm,height=1.5cm]{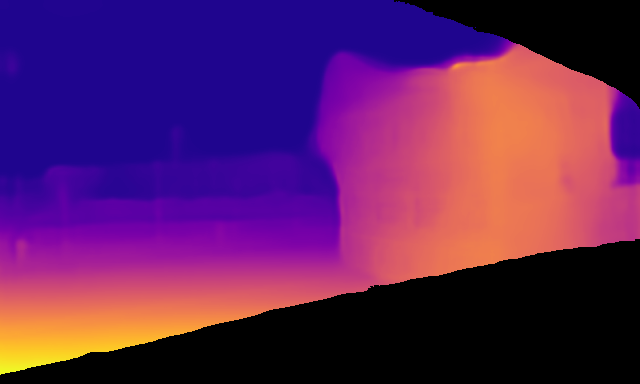}
\includegraphics[width=0.16\linewidth,height=1.5cm,height=1.5cm]{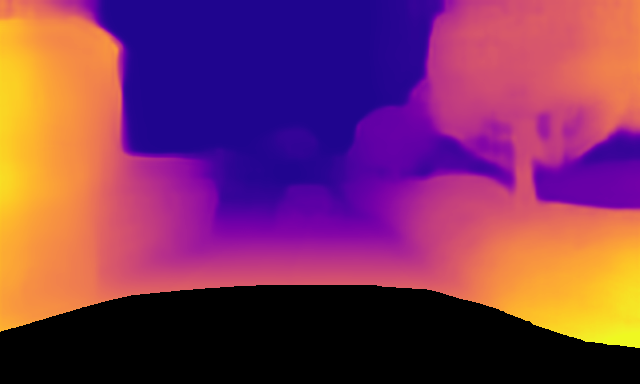}
}
\vspace{-4mm}
\\
\subfloat{
\includegraphics[width=0.16\linewidth,height=1.5cm,height=1.5cm]{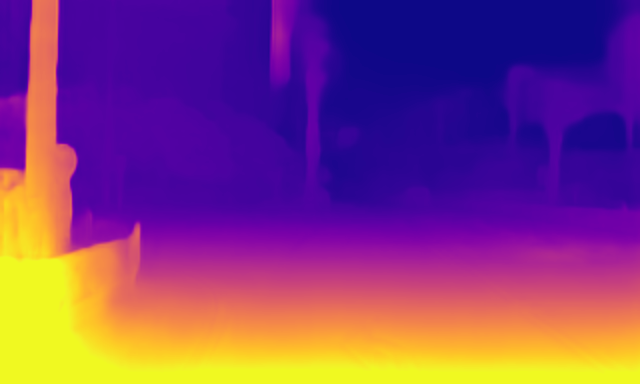}
\includegraphics[width=0.16\linewidth,height=1.5cm,height=1.5cm]{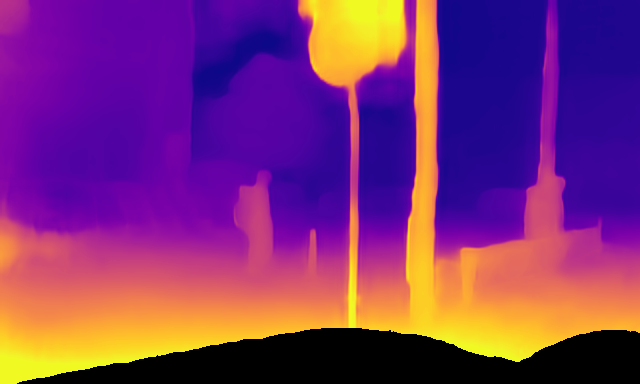}
\includegraphics[width=0.16\linewidth,height=1.5cm,height=1.5cm]{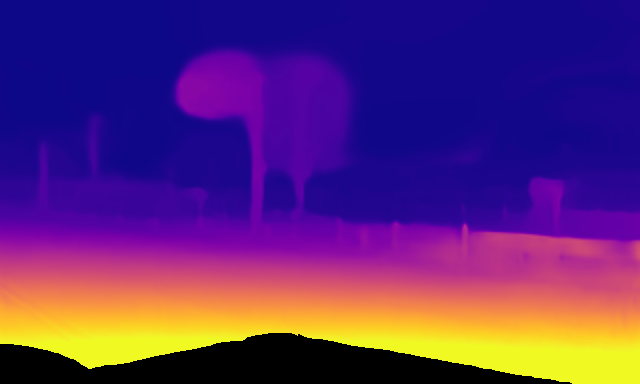}
\includegraphics[width=0.16\linewidth,height=1.5cm,height=1.5cm]{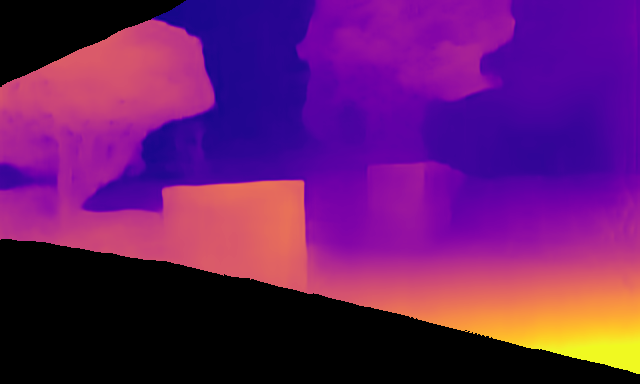}
\includegraphics[width=0.16\linewidth,height=1.5cm,height=1.5cm]{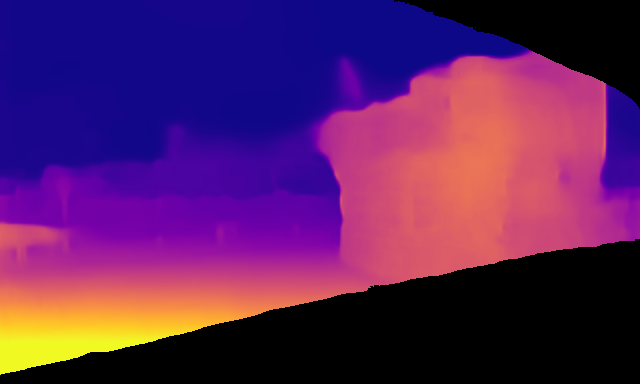}
\includegraphics[width=0.16\linewidth,height=1.5cm,height=1.5cm]{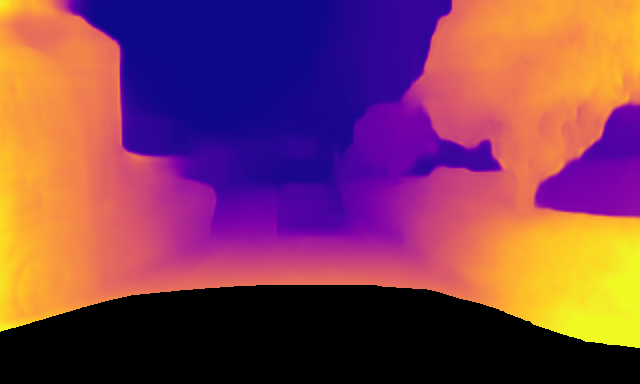}
}
\vspace{-4mm}

% \subfloat{
% \includegraphics[width=0.16\linewidth,height=1.5cm,height=1.5cm]{images/ddad_qualitative/ddad-val-lidar-camera_01-calc_normals_7_c7875beb.png}
% \includegraphics[width=0.16\linewidth,height=1.5cm,height=1.5cm]{images/ddad_qualitative/ddad-val-lidar-camera_05-calc_normals_7_11c2267c.png}
% \includegraphics[width=0.16\linewidth,height=1.5cm,height=1.5cm]{images/ddad_qualitative/ddad-val-lidar-camera_06-calc_normals_7_4d9ba474.png}
% \includegraphics[width=0.16\linewidth,height=1.5cm,height=1.5cm]{images/ddad_qualitative/ddad-val-lidar-camera_07-calc_normals_7_338b1655.png}
% \includegraphics[width=0.16\linewidth,height=1.5cm,height=1.5cm]{images/ddad_qualitative/ddad-val-lidar-camera_08-calc_normals_7_2ff0a179.png}
% \includegraphics[width=0.16\linewidth,height=1.5cm,height=1.5cm]{images/ddad_qualitative/ddad-val-lidar-camera_09-calc_normals_7_8df871f3.png}
% }
% \vspace{-2mm}
\caption{
\textbf{Qualitaitve results on the DDAD dataset.} The first row: RGB pictures of six cameras.  The second row: Mono2~\cite{godard2019digging}. The third row: FSM~\cite{9712255}. The fourth row: MCDP.
%using FSM on the \textit{DDAD} dataset.
}
% \vspace{-2mm}
%, including surface normals calculated from the predicted depth maps.}
\label{fig:ddad_qualitative}
\end{figure*}
%-------------------------------------------------------------------------------------------------------------------------------------------------

\begin{figure*}[th!]
\vspace{-2mm}
\centering
\subfloat{
\includegraphics[width=0.16\linewidth,height=1.5cm]{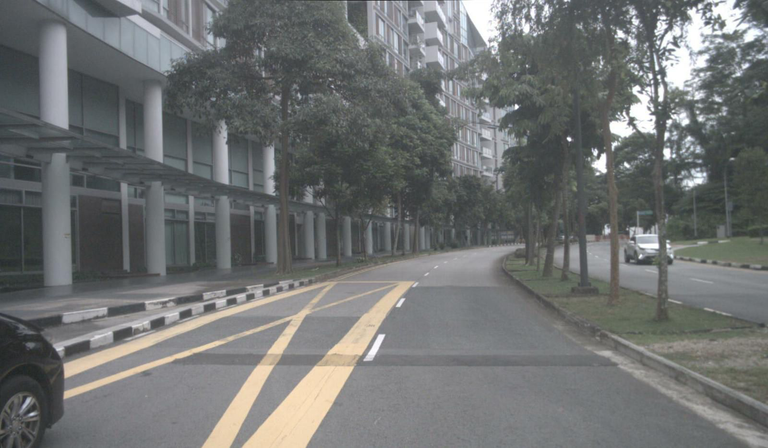}
\includegraphics[width=0.16\linewidth,height=1.5cm]{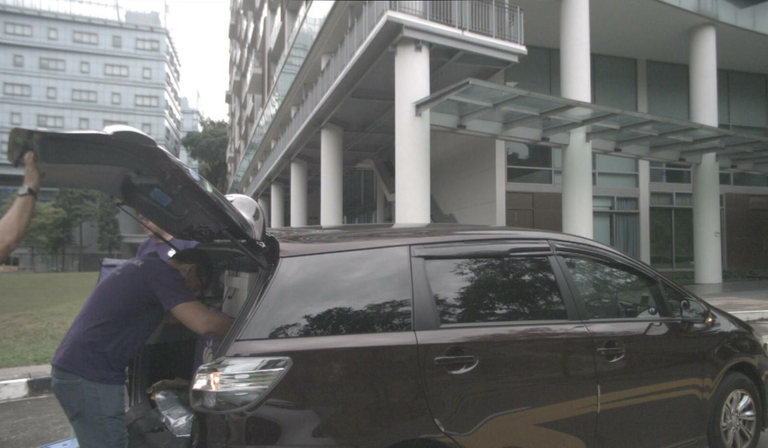}
\includegraphics[width=0.16\linewidth,height=1.5cm]{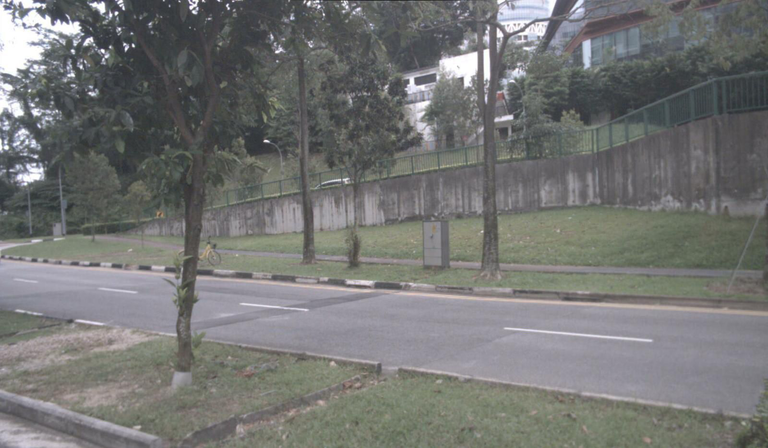}
\includegraphics[width=0.16\linewidth,height=1.5cm]{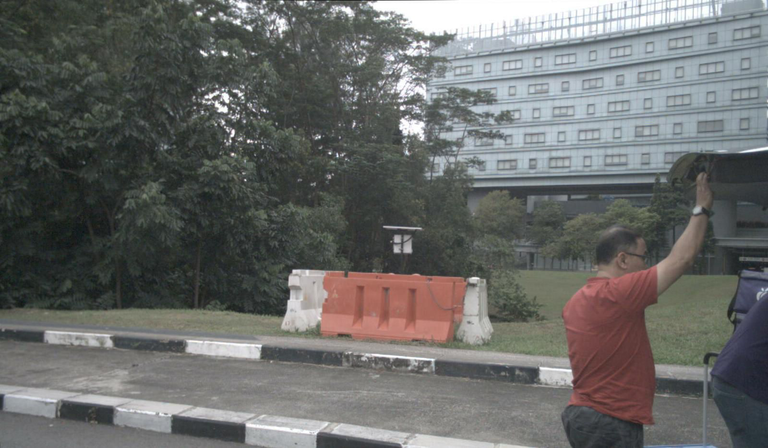}
\includegraphics[width=0.16\linewidth,height=1.5cm]{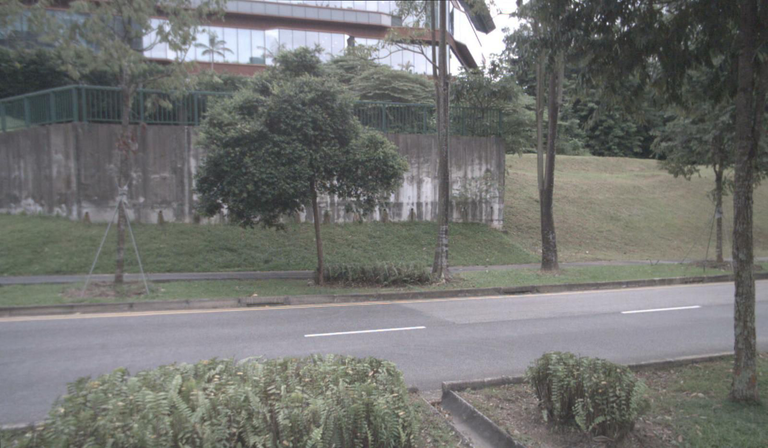}
\includegraphics[width=0.16\linewidth,height=1.5cm]{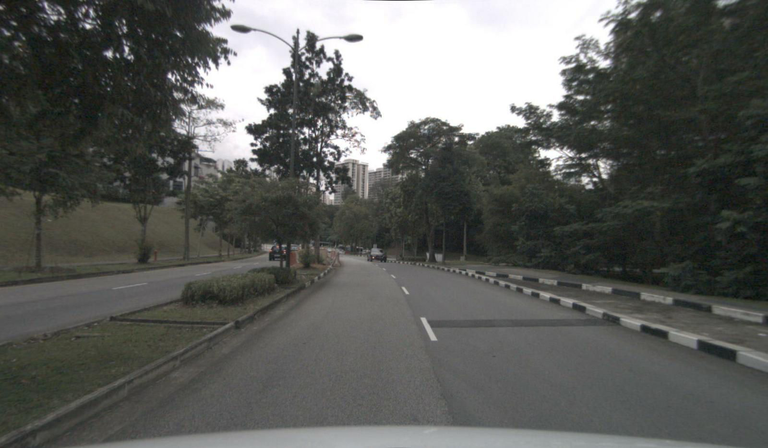}
}
\vspace{-4mm}
\\
\subfloat{
\includegraphics[width=0.16\linewidth,height=1.5cm,height=1.5cm]{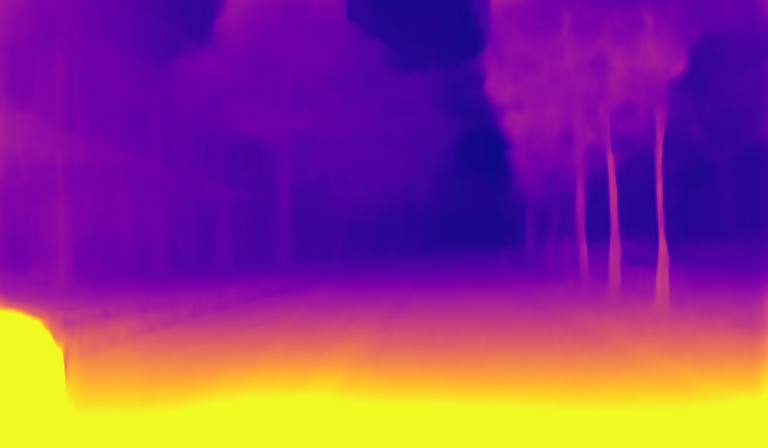}
\includegraphics[width=0.16\linewidth,height=1.5cm,height=1.5cm]{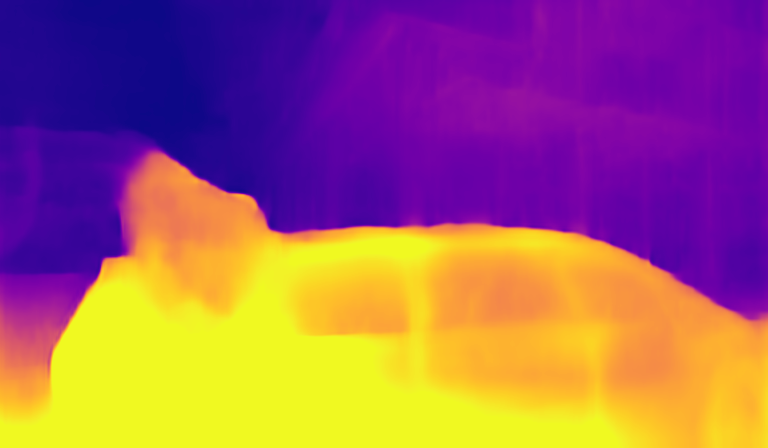}
\includegraphics[width=0.16\linewidth,height=1.5cm,height=1.5cm]{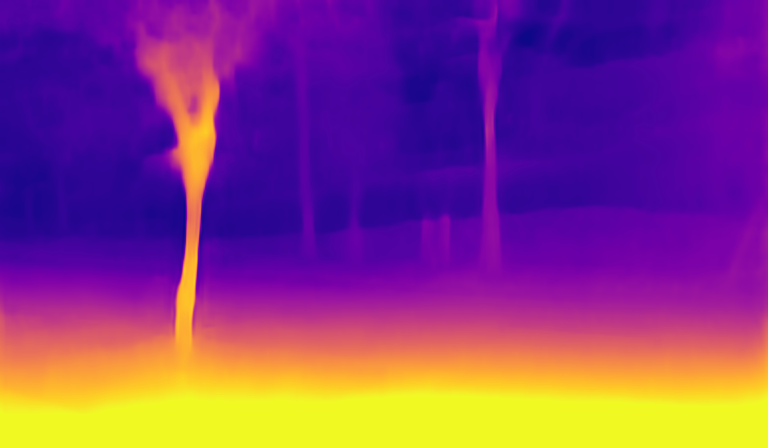}
\includegraphics[width=0.16\linewidth,height=1.5cm,height=1.5cm]{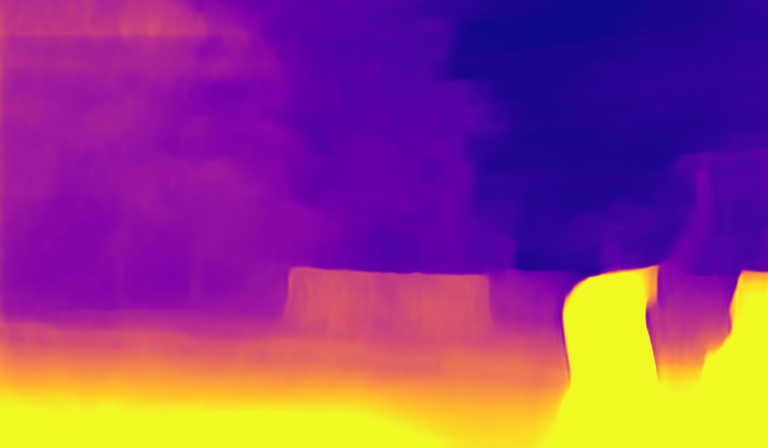}
\includegraphics[width=0.16\linewidth,height=1.5cm,height=1.5cm]{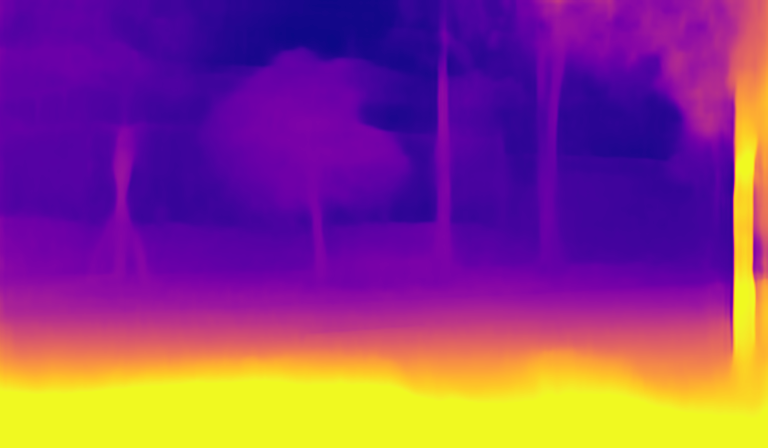}
\includegraphics[width=0.16\linewidth,height=1.5cm,height=1.5cm]{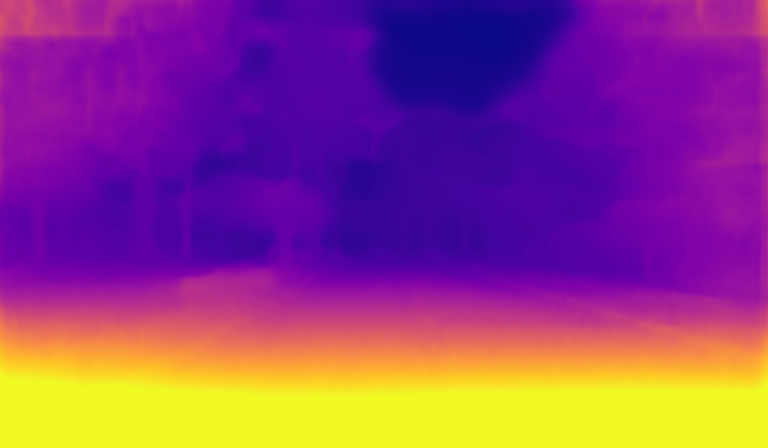}
}
\vspace{-4mm}
\\
\subfloat{
\includegraphics[width=0.16\linewidth,height=1.5cm,height=1.5cm]{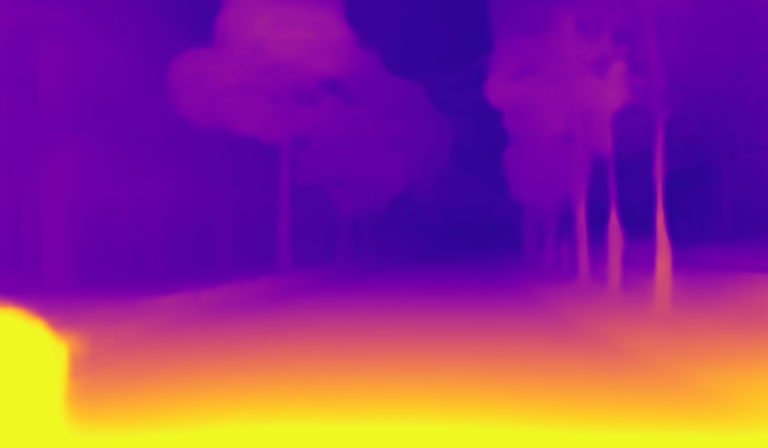}
\includegraphics[width=0.16\linewidth,height=1.5cm,height=1.5cm]{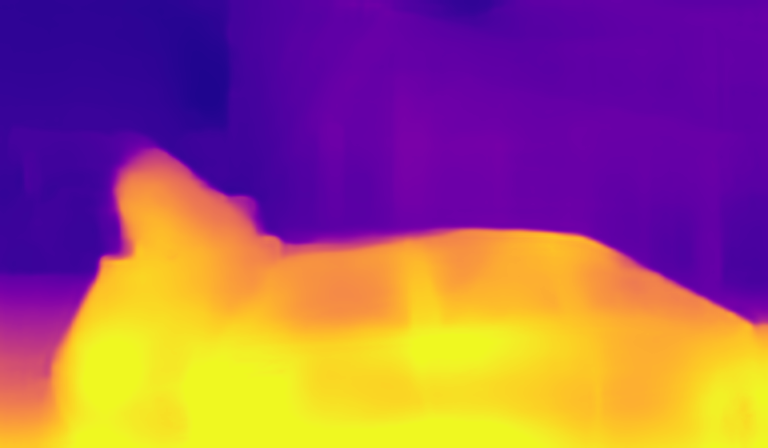}
\includegraphics[width=0.16\linewidth,height=1.5cm,height=1.5cm]{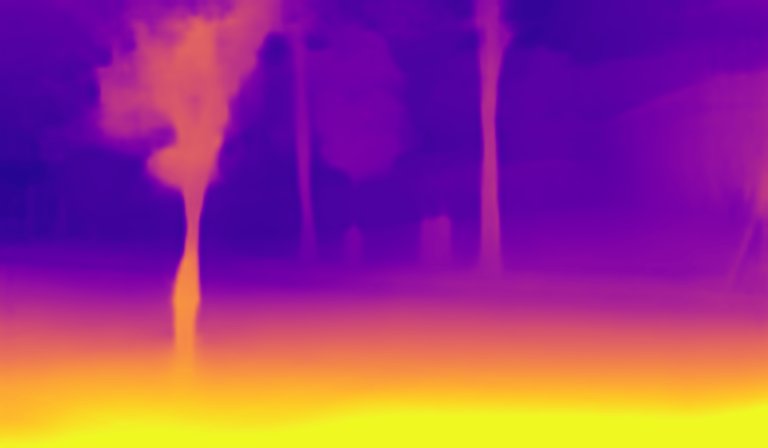}
\includegraphics[width=0.16\linewidth,height=1.5cm,height=1.5cm]{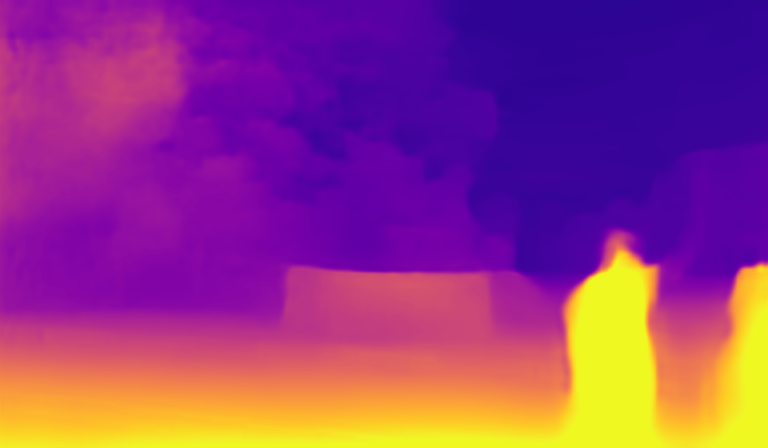}
\includegraphics[width=0.16\linewidth,height=1.5cm,height=1.5cm]{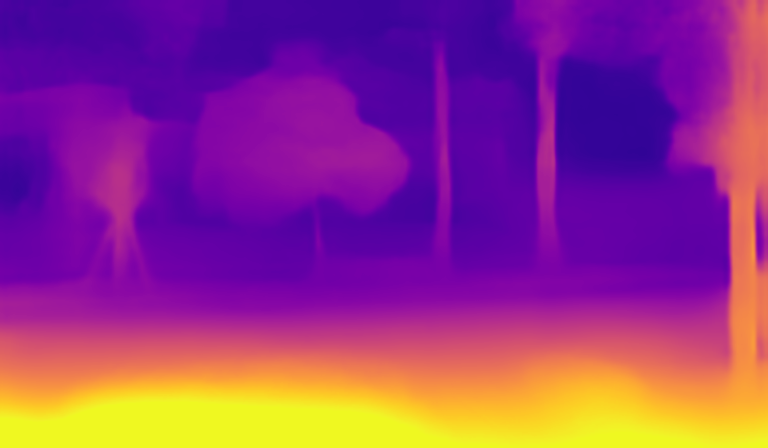}
\includegraphics[width=0.16\linewidth,height=1.5cm,height=1.5cm]{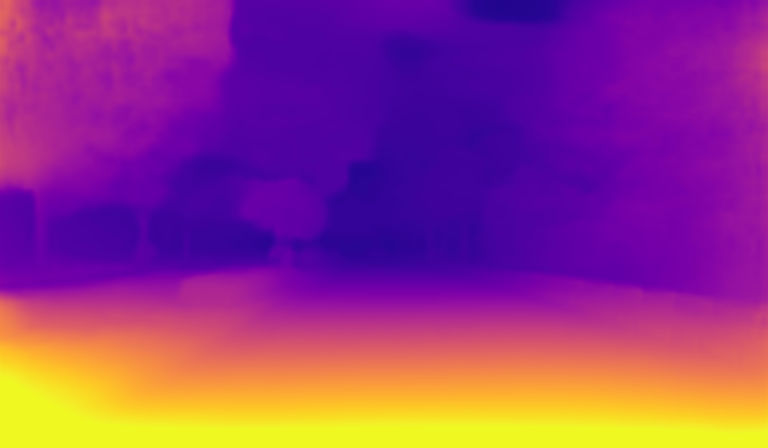}
}
\vspace{-4mm}
\\
\subfloat{
\includegraphics[width=0.16\linewidth,height=1.5cm,height=1.5cm]{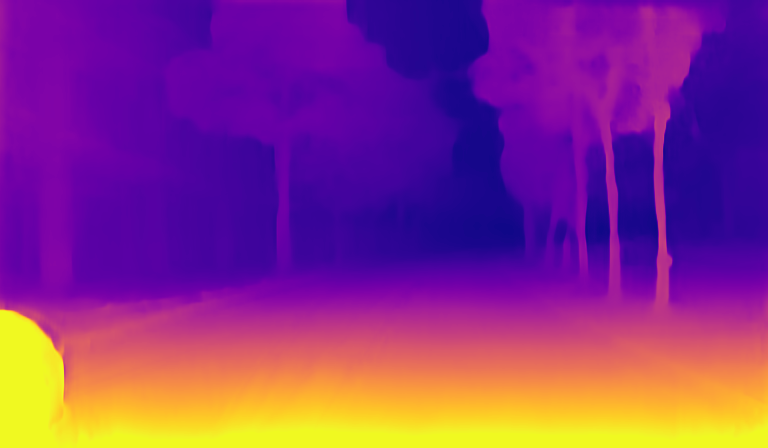}
\includegraphics[width=0.16\linewidth,height=1.5cm,height=1.5cm]{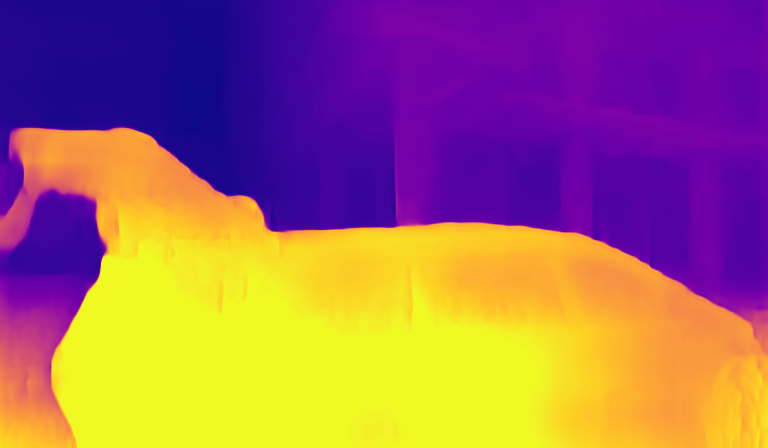}
\includegraphics[width=0.16\linewidth,height=1.5cm,height=1.5cm]{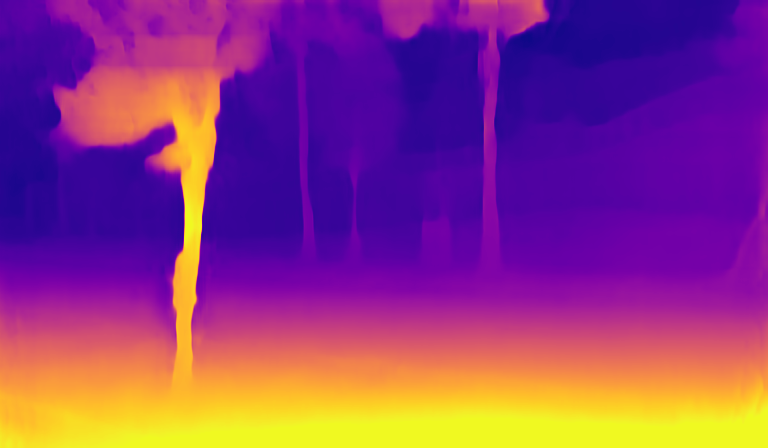}
\includegraphics[width=0.16\linewidth,height=1.5cm,height=1.5cm]{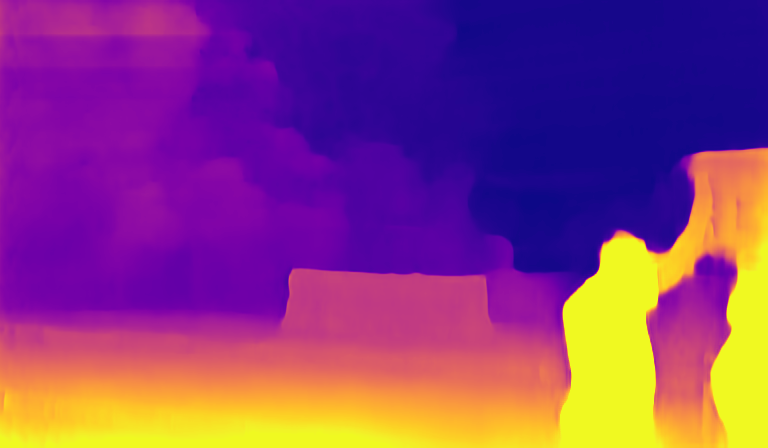}
\includegraphics[width=0.16\linewidth,height=1.5cm,height=1.5cm]{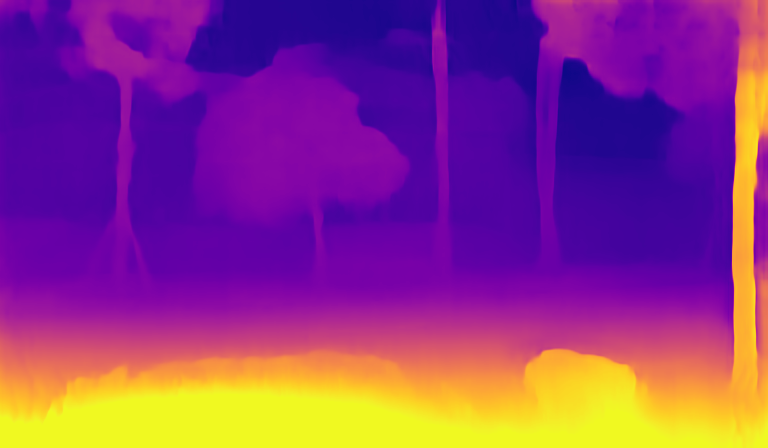}
\includegraphics[width=0.16\linewidth,height=1.5cm,height=1.5cm]{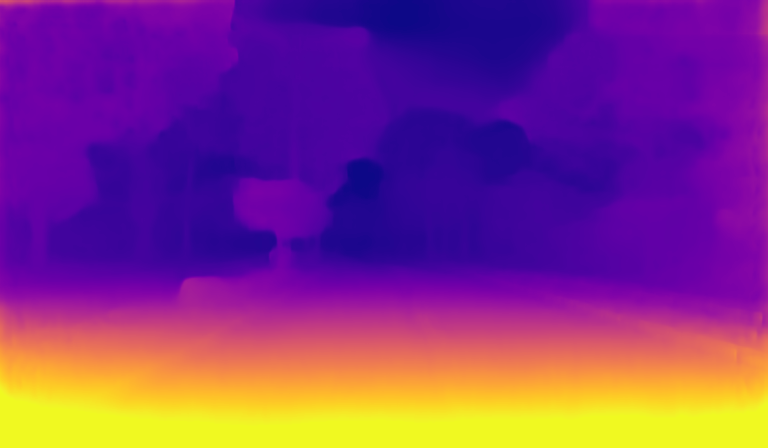}

}
\caption{
\textbf{Qualitaitve results on the NuScenes dataset.} The first row: RGB pictures of six cameras.  The second row: Mono2~\cite{godard2019digging}. The third row: FSM~\cite{9712255}. The fourth row: MCDP
}
% \vspace{-2mm}
%, including surface normals calculated from the predicted depth maps.}
\label{fig:nuScenes_qualitative}
\end{figure*}

\captionsetup[table]{skip=6pt}

%------------------------------------------------------------------------------------------------------------
\begin{table}%[t!]
% \vspace{-3mm}
%\vspace{-4mm}
%\rowcolors{2}{lightgray}{white}
%\renewcommand{\arraystretch}{}
\caption{
%\textbf{Quantitative depth evaluation of different methods on the DDAD \cite{guizilini20203d} dataset}, for distances up to 200m on the forward-facing camera. The symbol $^\star$ indicates a \emph{scale-aware} model, evaluated without median-scaling at test time.
Quantitative depth results on multi-camera datasets. All values are the average of six cameras. The distances are up to 200m in DDAD and 60m in Nuscenes. The symbol $^\star$ denotes sharing median-scaling at the test time. $m$ denotes the refine times. $M$ denotes the removal of masking.
}
%\begin{table}%
\centering
%\small
\setlength{\tabcolsep}{0.3em}
\subfloat[DDAD]{
\begin{tabular}{l|cccc}
\toprule
\textbf{Method}  &
Abs Rel$\downarrow$ &
Sq Rel$\downarrow$ &
RMSE$\downarrow$ &
$\delta_{1.25}$ $\uparrow$
\\
\toprule
Mono2 - M~\cite{godard2019digging} &
0.358 & 4.549 & 14.774 & 0.600 \\
Mono2 \cite{godard2019digging}&
0.220 & 4.235 & 14.125 & 0.751 \\
\midrule
PackNet~\cite{guizilini20203d}& 0.215 & 4.012 & 13.649 &  0.696\\
FSM$^\star$ ~\cite{9712255}  & 0.208 & 4.276 & 13.561 & 0.809 \\
FSM ~\cite{9712255} & 0.202 & 3.267 & 13.530 & 0.803 \\
SurroundDepth~\cite{wei2022surround} & 0.200 & 3.392 & 12.270 & 0.740 \\
\midrule
\textbf{MCDP} ($m$ = 0) & 0.217 & 3.832 & 13.451 & 0.767 \\
\textbf{MCDP} (w/o $L_{con}$, $m$ = 1) & 0.205 & 3.165 & 12.306 & 0.798 \\
\textbf{MCDP} ($m$ = 1 ) & 0.197 & 3.150 & 12.265 & 0.808 \\
\textbf{MCDP} ($m$ = 2 ) & \textbf{0.193} & \textbf{3.111} & \textbf{12.264} & \textbf{0.811}\\

\bottomrule
% \label{table:ddad depth}
\end{tabular}
}
\\
\vspace{-3mm}
\subfloat[NuScenes]{

%\end{table}

%------------------------------------------------------------------------------------------------------------
%\begin{table}[t!]
% \vspace{-3mm}
%\vspace{-4mm}
%\rowcolors{2}{lightgray}{white}

%\caption{
%\textbf{Quantitative depth evaluation of different methods on the NuScenes \cite{caesar2020nuscenes} dataset}. All values are the average of six cameras, for distances up to 80m. The symbol $^\star$ denotes sharing median-scaling at the test time. $m$ denotes the refine times.}
%\centering

%\setlength{\tabcolsep}{0.3em}
\begin{tabular}{l|cccc}
\toprule
\textbf{Method}  &
Abs Rel$\downarrow$ &
Sq Rel$\downarrow$ &
RMSE$\downarrow$ &
$\delta_{1.25}$ $\uparrow$
\\
\toprule
Mono2 & 0.309 & 3.453 & 8.012 & 0.663 \\
\midrule
PackNet~\cite{guizilini20203d}& 0.303 & 3.154 & 7.014 & 0.655\\
FSM$^\star$ ~\cite{9712255}  & 0.275 & 3.182 & 6.908 & 0.682 \\
FSM ~\cite{9712255} & 0.270 & 3.185 & 6.832 & 6.689 \\
SurroundDepth~\cite{wei2022surround} &0.245 & 3.067  &6.835 &  \textbf{0.719}\\
\midrule
\textbf{MCDP} ($m$ = 0) & 0.296 & 3.076 & 7.578 & 0.672 \\
\textbf{MCDP} (w/o $L_{con}$, $m$ = 1) & 0.257 & 3.052 & 6.849 & 0.702 \\
\textbf{MCDP} ($m$ = 1 ) & 0.240 & 3.031 & 6.825 & 0.716 \\
\textbf{MCDP} ($m$ = 2 ) & \textbf{0.237} & \textbf{3.030} & \textbf{6.822} &\textbf{0.719} \\
\bottomrule
\end{tabular}

}
\vspace{-3mm}
\label{table:both datasets depth result}
\end{table}

%------------------------------------------------------------------------------------------------------------
\iffalse
\begin{table}%
\caption{\textbf{Depth estimation results of each camera on NuScenes datasets}. The symbol $^\star$ denotes  shared median-scaling and $m$ denotes the refine times.}
\small
% \vspace{-1mm}
\centering
\renewcommand{\arraystretch}{1.05}
\setlength{\tabcolsep}{0.29em}
%\subfloat[]{
\begin{tabular}{l|cccccc}
\toprule
\multirow{2}{*}{\textbf{Method}} &
\multicolumn{6}{c}{Abs Rel $\downarrow$} \\
\cmidrule{2-7}
& \textit{Front} & \textit{F.Left} & \textit{F.Right} & \textit{B.Left} & \textit{B.Right} & \textit{Back}\\

\midrule
Mono2$^\star$ & 0.201 &  0.272 &  0.390 &  0.231 &  0.401 & 0.275   \\
Mono2 & 0.196 & 0.286 & 0.370 & 0.296 & 0.42 & 0.286  \\

\midrule
FSM$^\star$ & 0.165 & 0.260 & 0.363 & 0.279 & 0.382 & 0.201  \\
FSM & 0.164 &  0.262 &  0.341 &  0.269 &  0.390 &  0.194   \\

\midrule
\textbf{MCDP ($m$ = 0)} & 0.189 & 0.318 & 0.377 & 0.272 &0.377 & 0.245   \\
\textbf{MCDP (w/o $L_{con}$, $m$ = 1)} &0.181 &0.238 & 0.329 & 0.260 & 0.329 & 0.207 \\
\textbf{MCDP ($m$ = 2)}  &\textbf{0.160} & \textbf{0.226} & \textbf{0.292} & \textbf{0.247} & \textbf{0.310} & \textbf{0.189} \\

\bottomrule
\end{tabular}
%}

\label{tab:each camera results}
\vspace{-3mm}
\end{table}
\fi
%------------------------------------------------------------------------------------------------------------

\begin{table}[ht]%
\caption{\textbf{Depth consistency results of each camera on two datasets}.}
% \vspace{-1mm}
\centering
\renewcommand{\arraystretch}{1.05}
\setlength{\tabcolsep}{0.22em}
%\subfloat[DDAD]{
\begin{tabular}{l|l|cccccc}
\toprule
\multirow{2}{*}{\textbf{Dataset}} & \multirow{2}{*}{\textbf{Method}} &
\multicolumn{6}{c}{Dep Con$\downarrow$} \\
%\cmidrule{3-8}&
\cline{3-8}
& & \textit{Front} & \textit{F.Left} & \textit{F.Right} & \textit{B.Left} & \textit{B.Right} & \textit{Back} \\

\midrule
\multirow{2}{*}{\textbf{DDAD}}& Mono2 & 0.637    & 0.279     & 0.330    & 0.317     & 0.336     & 0.391    \\
&MCDP &  \textbf{0.621}   & \textbf{0.268}    & \textbf{0.205}    & \textbf{0.228}     & \textbf{0.292}     & \textbf{0.330}  \\

\midrule
\multirow{2}{*}{\textbf{NuScenes}} & Mono2 & 0.279   & 0.337     & 0.316    & 0.354 & 0.277  & 0.351\\
&MCDP  & \textbf{0.255}    &\textbf{0.318}    & \textbf{0.244}   & \textbf{0.301}   & \textbf{0.240}    & \textbf{0.337}   \\

\bottomrule
\end{tabular}
%}

\label{tab:Dep con result}
\vspace{-3mm}
\end{table}
%------------------------------------------------------------------------------------------------------------

\begin{table}[ht]%
\caption{\textbf{Quantitative depth results on KITTI dataset. $m$ denotes the refine times. %Monodepth is the baseline for our method, which do not used multi-camera collaborative depth estimation refinement and depth consistency loss $L_{con}$.
}}
% \vspace{-1mm}
\centering
\renewcommand{\arraystretch}{1.05}
\setlength{\tabcolsep}{0.22em}
%\subfloat[DDAD]{
\begin{tabular}{l|l|cccc}
\toprule
\textbf{Method} & $m$ & Abs Rel$\downarrow$ & Sq Rel $\downarrow$& RMSE$\downarrow$ &$\delta_{1.25}$ $\uparrow$ \\
\midrule
Mono2~\cite{godard2019digging} & - & 0.106 & 0.806 & 4.630 & 0.876 \\
%MCDP      & 0 & 0.105 & 0.803 & 4.619 & 0.876 \\
MCDP (w/o $L_{con}$) & 1 & 0.105 & 0.805 & 4.622 & 0.878\\
%MCDP      & 1 & \textbf{0.102} &\textbf{ 0.798} & 4.620 & \textbf{0.880} \\
MCDP      & 2 &\textbf{ 0.102} & \textbf{0.798} & \textbf{4.619}&\textbf{ 0.880}\\

\bottomrule
\end{tabular}
%}

\label{tab:KTIIT result}
\vspace{-3mm}
\end{table}
%------------------------------------------------------------------------------------------------------------

\subsection{Implementation details}
We implement the proposed scheme in PyTorch~\cite{paszke2019pytorch}, training networks for 20 epochs on two Nvidia RTX 3090 GPU. The batch size is set to 12 in DDAD dataset and 8 in NuScenes dataset. We jointly train the depth network, pose network and weight network with Adam Optimizer~\cite{kingma2014adam} with $\beta_1$ = 0.9, $\beta_2$ = 0.999. The initial learning rate is set to 1$e^{-4}$ and decay for the first 15 epochs which is then dropped to 1$e^{-5}$ for the remainder.  We set the SSIM weight to $\alpha$ = 0.85, the weight of smoothness term to 0.001, the weight of depth consistency loss to $\lambda$ = 0.001 and the number of depth basis $n$ = 32. Training takes 40 hours for DDAD dataset and 140 hours for NuScenes dataset. In the testing phase, we do not flip the image left and right and take the average, although this is used in many works~\cite{xu2021weakly,bhat2021adabins} to improve the accuracy.

\paragraph{Depth Estimation Network} Following the Mono2~\cite{godard2019digging},  we use the ResNet18-based depth and pose networks for depth estimation. As the underlying network for monocular self-supervised estimation, the details are consistent with Mono2 and remain unchanged in all our experiments.

\paragraph{Self-Occlusions}
As shown in Fig.~\ref{fig:mask}, part of the images is obscured by the car itself, and the occluded area in multiple positions is very large. This occlusion has a serious impact on the calculation of the photometric loss. 
In addition, there are also self-occlusion in the overlap of multi-camera, which will also bring wrong guidance to the refinement work. Referring to ~\cite{9712255} for dealing with occlusion, we manually draw the mask for each camera, which only needs to be drawn once for the entire dataset. In the calculation of $L_p$, $L_{con}$ and $\hat{F}$, masks are used to remove the occluded parts, which can remove erroneous information for the network in advance.

%\vspace{-0.8em}
\subsection{Depth Estimation Performance}

\paragraph{Refinement by Depth Basis.}
In the two datasets, we mix the pictures of all cameras together and train them by the same model. We first evaluate the overall impact of different methods on six cameras. As shown in Tab.~\ref{table:both datasets depth result}, we take Mono2~\cite{godard2019digging} as the benchmark, and the effect of mask is obvious. As described in \ref{sec:Multi-camera framework}, MCDP can greatly improve the performance of all metrics. With only one refine adjustment, the \emph{Abs Rel} reduces from 0.217 to 0.197 (10.1\%) on DDAD and 0.296 to 0.240  (23.3\%) on NuScenes, and $\delta_{1.25}$ can improve from 0.808 to 0.767 (5.0\%) on DDAD and from 0.672 to 0.716 (6.1\%) on NuScenes. This proves that the MCDP can use the overlapping part to improve the overall prediction performance of the cameras.

Furthermore, the characteristic of multi-camera framework is that it can be optimized many times. We refine it twice, and the performance of the model has been further improved. The \emph{abs rel} reduces from 0.197 to 0.193 on DDAD and from 0.237 to 0.240 on NuScenes. Another metric $\delta_{1.25}$  improves from 0.808 to 0.811 on DDAD and from 0.716 to 0.719 on NuScenes. This proves that multiple refinement iterations can further improve the model performance. In Fig.~\ref{fig:ddad_qualitative} and Fig.~\ref{fig:nuScenes_qualitative}, we show the quantitative results separately in DDAD and NuScenes. Compared with Mono2 and FSM, our method can significantly improve the accuracy of object edge and overlapping area.

\paragraph{Assessment  for Depth Consistency Loss.} In the Tab.~\ref{table:both datasets depth result}, we compare the effects before and after using $L_{con}$ in two datasets. Compared with the benchmark, \emph{Abs Rel} can reduce from 0.220 to 0.217 (1.3\%) in DDAD and from 0.303 to 0.296 (2.6\%) with $L_{con}$. More importantly, $L_{con}$ can be used in refinement to improve the performance of the model. In the MCDP scheme, we show the results before and after using $L_{con}$ based on refining once, and \emph{Abs Rel} reduces from 0.205 to 0.197 (4.0\%) in DDAD and 0.257 to 0.240 (7.0\%) in NuScenes. This proves that the $L_{con}$ can improve the performance of the model by constraining the depth of the overlap between cameras. Combined with our results, the MCDP can  iteratively refine the estimated depth for many times, and generates the state-of-the-art models.

\iffalse
\subsection{Assessment for Each Camera}
We also evaluate the effect of our scheme on each camera. In Tab.~\ref{tab:each camera results}, we show the \emph{Abs Rel} of each camera in different models. Due to the different viewing angles of each camera, the depth distribution of each camera is not the same. This is a great challenge to the model and leads to great differences in the prediction results between the cameras. MCDP also achieves the best performance in each camera.
\fi

\subsection{Depth Consistensy Metric}
The depth consistency in overlapping parts is our focus. We use Equation ~\ref{eq:Dep Con} to evaluate the depth consistency error of overlapping parts. As shown in Tab.~\ref{tab:Dep con result}, due to the different depth distribution caused by the camera position, the depth consistency between cameras varies greatly. Left and right view cameras have better consistency than front and back cameras. Compared with the benchmark, MCDP can effectively improve the depth consistency between cameras. This is verified on six cameras in the both datasets.

%--------------------------------------------------------------------------------
\iffalse
\begin{figure}[t]
  \begin{center}
    \small
    \hspace{4mm} Basis 1 \hspace{14mm} Basis 2 \hspace{14mm} Basis 3 \\
    \rotatebox{90}{\hspace{3.5mm} Scene 2 \hspace{6.5mm} Scene 1}
    \includegraphics[width=0.9\linewidth]{figures/basis/basis.png}
  \end{center}
  \caption{
    Visualisation of the influence of the basis on depth refinement.
    Compared to the refined depth map, a specific depth basis is used to colourise the input image (red and blue are positive and negative values, respectively).
    Columns represent basis entries (1-3).
    Rows represent two different input scenes.
  }
  \label{fig:depth basis}
  %\vspace{2mm}\hrule
\end{figure}
 \fi
%--------------------------------------------------------------------------------
\iffalse
\subsection{Depth Basis Visualise}
In Figure.~\ref{fig:depth basis} we visualise the depth basis on the color image. An interesting observation is that the depth basis seem to correspond to the specific image feature, such as edge profile, relative distance of objects. This also explains why depth results can be improved by adjusting the linear combination of depth basis. We also compare the regions of influence for two different scenes and can observe a certain degree of consistency.
 \fi

\subsection{Assessment for Robustness}
In order to verify the robustness of our method, we select more diverse data for testing. KITTI~\cite{geiger2012we} is a stereo camera dataset, and it is not suitable for the setting of small overlapping areas between cameras that we are concerned about. 
%However, we can also evaluate our method on KITTI to show the robustness.
As shown in Tab.~\ref{tab:KTIIT result}, our method is also effective on KITTI. 
%Compared with monocular depth prediction, depth basis for consistent structure estimation and $L_{con}$ in our scheme both improve the performance. 
%But the improvement effect is not as obvious as that on DDAD and NuScenes datasets. This is because KITTI is a stereo camera dataset, and the images captured by the two cameras are very similar in large overlapping areas, which makes the additional information obtained from the other camera very limited during the multi-camera collaborative optimization process.
It should be noted that, compared to the other multi-camera datasets like DDAD and NuScenes, the biggest advantage of KITTI is its rectified stereo cameras, which can calculate the disparity between images. However, it is impossible with two cameras placed arbitrarily to compute disparity, and our universal approach cannot take advantage of this particular advantage. %Therefore, it is not suitable for predicting accurate depth maps with our scheme, and 
Our approach also shows good robustness with large overlap on KITTI.

\section{Conclusion}
%Using 3D perception technology to predict the geometric state of the real scene from a series of images is a very exciting computer vision technology. 
%In the context of self supervised learning
In this paper, we extend monocular depth prediction to multi-camera depth prediction at any position. In extreme scenes where the overlapping area between cameras is very small, we propose a multi-camera joint depth estimation scheme. We introduce a series of key technologies to improve the effect of depth prediction: depth basis for consistent structure estimation  allows the model to use overlapping area information in inference, and the depth consistency loss restricts the depth consistency among multi-camera. We demonstrate the capabilities of my approach and their parametric performance in two standard autonomous driving datasets. 
%In the future work, we plan to improve the ability of multi-camera to estimate dynamic objects in the context of self-supervised learning.

\section{Acknowledgments}
This work was supported by National Natural Science Foundation of China under Grants 61922027, 6207115 and 61932022, and in part by National Key Research and Development Project under Grant 2019YFE0109600.

%\begin{acks}
%To Robert, for the bagels and explaining CMYK and color spaces.
%\end{acks}

\bibliographystyle{ACM-Reference-Format}
\bibliography{sample-base}

\appendix

\end{document}